\documentclass[pdflatex,sn-mathphys-num]{sn-jnl}

\usepackage{graphicx}
\usepackage{array}
\usepackage{amssymb}
\usepackage{amsmath}
\usepackage{stfloats}
\usepackage{tabularx,booktabs}
\usepackage{hyperref}
\usepackage{floatrow}
\usepackage{multirow}
\usepackage{threeparttable}
\usepackage{makecell}
\usepackage{multirow}
\usepackage{bm}
\usepackage{caption}
\usepackage{xcolor}
\usepackage{booktabs}
\usepackage{setspace}
\usepackage{array}
\usepackage[labelfont=bf]{caption}
\usepackage{graphicx}
\usepackage{newfloat}
\usepackage{hhline}
\usepackage{listings}
\usepackage{tikz}
\usepackage{verbatim}
\usepackage{latexsym}
\usepackage[T1]{fontenc}
\usepackage[utf8]{inputenc}
\usepackage{microtype}
\usepackage{inconsolata}
\usepackage{subfigure}
\usepackage{multirow}
\usepackage{algorithmic}
\usepackage{algorithm}
\usepackage{amsfonts}
\usepackage{textcomp}
\usepackage{url}
\usepackage{makecell}
\usepackage{ragged2e}
\usepackage{pifont}
\usepackage{amssymb}
\usepackage{xcolor} 
\usepackage{arydshln}  
\usepackage{array}
\usepackage[normalem]{ulem} 
\usepackage[numbers,sort&compress]{natbib}

\usepackage{color}
\usepackage{colortbl}
\definecolor{bc1}{RGB}{222,234,246}
\definecolor{bc2}{RGB}{251,228,213}
\definecolor{bc3}{RGB}{226,239,217}
\definecolor{bc4}{RGB}{230,193,222}
\definecolor{backgroud}{RGB}{240,240,240}
\usepackage{CJKutf8}

\usepackage{CJKutf8}
\usepackage{lineno}
\usepackage[most]{tcolorbox}

\theoremstyle{thmstyleone}%
\theoremstyle{thmstyletwo}%

\theoremstyle{thmstylethree}%

\begin{document}

\title{LASQ: A \underline{L}ow-resource \underline{A}spect-based \underline{S}entiment \underline{Q}uadruple Extraction Dataset }


\author[1 2]{\fnm{Aizihaierjiang} \sur{Yusufu}}\email{azhar520abida@gmail.com}

\author[2]{\fnm{Jiang} \sur{Liu}}\email{wugangliujiang@gmail.com)}\equalcont{These authors contributed equally to this work.}

\author[3]{\fnm{Kamran} \sur{Aziz}}\email{kamran.aziz@hainan-biuh.edu.cn}

\author[4]{\fnm{Abidan} \sur{Ainiwaer}}\email{abidan@xju.edu.cn}

\author[2]{\fnm{Bobo} \sur{Li}}\email{libobo@nus.edu.sg}

\author[2]{\fnm{Fei} \sur{Li}}\email{foxlf823@gmail.com}

\author[2]{\fnm{Donghong} \sur{Ji}}\email{dhji@whu.edu.cn}

\author*[1]{\fnm{Aizierguli} \sur{Yusufu}}\email{Azragul2010@126.com}

\affil*[1]{\orgdiv{Xinjiang Engineering Technology Research Center for Smart Education and Application, School of Computer Science and Technology}, \orgname{Xinjiang Normal University}, \orgaddress{ \city{Urumqi}, \postcode{830054}, \state{Xinjiang}, \country{China}}}

\affil[2]{\orgdiv{Key Laboratory of Aerospace Information Security and Trusted Computing, School of Cyber Science and Engineering}, \orgname{Wuhan University}, \orgaddress{\city{Wuhan}, \postcode{430000},\state{Hubei}, \country{China}}}

\affil[3]{\orgname{Hainan Bielefeld University of Applied Sciences}, \orgaddress{\street{Central Avenue and East 13}, \city{Danzhou}, \postcode{578001}, \state{Hainan}, \country{China}}}

\affil[4]{\orgname{Journalism \& Communication College, Xinjiang University}, \orgaddress{\street{No. 666 Shengli Road}, \city{Urumqi}, \postcode{830046}, \state{Xinjiang}, \country{China}}}


\abstract{In recent years, aspect-based sentiment analysis (ABSA) has made rapid progress and shown strong practical value. However, existing research and benchmarks are largely concentrated on high-resource languages, leaving fine-grained sentiment extraction in low-resource languages under-explored. To address this gap, we constructed the first \textbf{L}ow-resource languages \textbf{A}spect-based \textbf{S}entiment \textbf{Q}uadruple dataset, named LASQ, which includes two low-resource languages: Uzbek and Uyghur. Secondly, it includes a fine-grained target-aspect-opinion-sentiment quadruple extraction task. To facilitate future research, we designed a grid-tagging model that integrates syntactic knowledge. This model incorporates part-of-speech (POS) and dependency knowledge into the model through our designed \textbf{S}yntax \textbf{K}nowledge \textbf{E}mbedding \textbf{M}odule (SKEM), thereby alleviating the lexical sparsity problem caused by agglutinative languages. Experiments on LASQ demonstrate consistent gains over competitive baselines, validating both the dataset’s utility and the effectiveness of the proposed modeling approach. }




\maketitle

\section{Introduction}\label{sec1}

Aspect-based sentiment analysis (ABSA) aims to identify the specific aspects mentioned in text and determine their associated sentiment polarity, offering a more fine-grained and interpretable paradigm for opinion understanding than traditional document- or sentence-level sentiment classification~\citep{ABSA1, ABSA2}. Currently, researchers have proposed many ABSA datasets, and we have summarized the characteristics of these datasets, as shown in Table \ref{tab:existing_ABSA_datasets}. For ABSA datasets with high-resource languages, from ASTE \citep{aste} and MAMS \citep{mams} to ACOS \citep{ACOS}, the focus has shifted from extracting aspect-opinion-sentiment triples to extracting more granular target-aspect-opinion-sentiment quadruples. However, they still only focus on English. Later, the proposal of DiaASQ \citep{meatwp} extended the monolingual scope to multilingual, including not only English but also Chinese.

\begin{table}[!h]
\fontsize{12}{14}\selectfont
\setlength{\tabcolsep}{0.8mm}
\setlength{\arrayrulewidth}{0.1mm}
\centering
\resizebox{0.85\textwidth}{!}{
\begin{tabular}{l c c c c c c c c c c c c}
\hline
& \phantom{}&Target&\phantom{}&Aspect&\phantom{}&Opinion&\phantom{}&Sentiment&\phantom{}&Multilingual&\phantom{}&Low-resource\\
\hline
ASTE&\phantom{}&\textcolor{red}{\ding{55}}&\phantom{}&\textcolor{green}{\ding{51}}&\phantom{}&\textcolor{green}{\ding{51}}&\phantom{}&\textcolor{green}{\ding{51}}&\phantom{}&\textcolor{red}{\ding{55}}&\phantom{}&\textcolor{red}{\ding{55}}\\
MAMS&\phantom{}&\textcolor{red}{\ding{55}}&\phantom{}&\textcolor{green}{\ding{51}}&\phantom{}&\textcolor{green}{\ding{51}}&\phantom{}&\textcolor{green}{\ding{51}}&\phantom{}&\textcolor{red}{\ding{55}}&\phantom{}&\textcolor{red}{\ding{55}}\\
ACOS&\phantom{}&\textcolor{green}{\ding{51}}&\phantom{}&\textcolor{green}{\ding{51}}&\phantom{}&\textcolor{green}{\ding{51}}&\phantom{}&\textcolor{green}{\ding{51}}&\phantom{}&\textcolor{red}{\ding{55}}&\phantom{}&\textcolor{red}{\ding{55}}\\
DiaASQ&\phantom{}&\textcolor{green}{\ding{51}}&\phantom{}&\textcolor{green}{\ding{51}}&\phantom{}&\textcolor{green}{\ding{51}}&\phantom{}&\textcolor{green}{\ding{51}}&\phantom{}&\textcolor{green}{\ding{51}}&\phantom{}&\textcolor{red}{\ding{55}}\\
\hdashline
Uzbek&\phantom{}&\textcolor{red}{\ding{55}}&\phantom{}&\textcolor{green}{\ding{51}}&\phantom{}&\textcolor{red}{\ding{55}}&\phantom{}&\textcolor{green}{\ding{51}}&\phantom{}&\textcolor{red}{\ding{55}}&\phantom{}&\textcolor{green}{\ding{51}}\\
Vietnamese&\phantom{}&\textcolor{red}{\ding{55}}&\phantom{}&\textcolor{green}{\ding{51}}&\phantom{}&\textcolor{red}{\ding{55}}&\phantom{}&\textcolor{green}{\ding{51}}&\phantom{}&\textcolor{red}{\ding{55}}&\phantom{}&\textcolor{green}{\ding{51}}\\
Telugu&\phantom{}&\textcolor{red}{\ding{55}}&\phantom{}&\textcolor{green}{\ding{51}}&\phantom{}&\textcolor{red}{\ding{55}}&\phantom{}&\textcolor{green}{\ding{51}}&\phantom{}&\textcolor{red}{\ding{55}}&\phantom{}&\textcolor{green}{\ding{51}}\\
Urdu&\phantom{}&\textcolor{red}{\ding{55}}&\phantom{}&\textcolor{green}{\ding{51}}&\phantom{}&\textcolor{red}{\ding{55}}&\phantom{}&\textcolor{green}{\ding{51}}&\phantom{}&\textcolor{red}{\ding{55}}&\phantom{}&\textcolor{green}{\ding{51}}\\
\hdashline
Our&\phantom{}&\textcolor{green}{\ding{51}}&\phantom{}&\textcolor{green}{\ding{51}}&\phantom{}&\textcolor{green}{\ding{51}}&\phantom{}&\textcolor{green}{\ding{51}}&\phantom{}&\textcolor{green}{\ding{51}}&\phantom{}&\textcolor{green}{\ding{51}}\\
\hline
\end{tabular}
}
\caption{Comparison between our dataset and existing ABSA datasets.}
\label{tab:existing_ABSA_datasets}
\end{table}

The development of high-resource languages is crucial, but the development of low-resource languages is equally important, as residents of most countries in the world tend to use their native languages. Recently, some ABSA datasets for low-resource languages have begun to emerge, including Uzbek \citep{uzabsa}, Vietnamese \citep{Vietnamese}, Telugu \citep{Telugu}, and Urdu \citep{Urdu}. However, these datasets still have two shortcomings. One is that they only focus on aspect-sentiment pair extraction, and the other is that they only contain a single low-resource language.

\begin{figure}[h]
\centering
\includegraphics[width=0.60\textwidth]{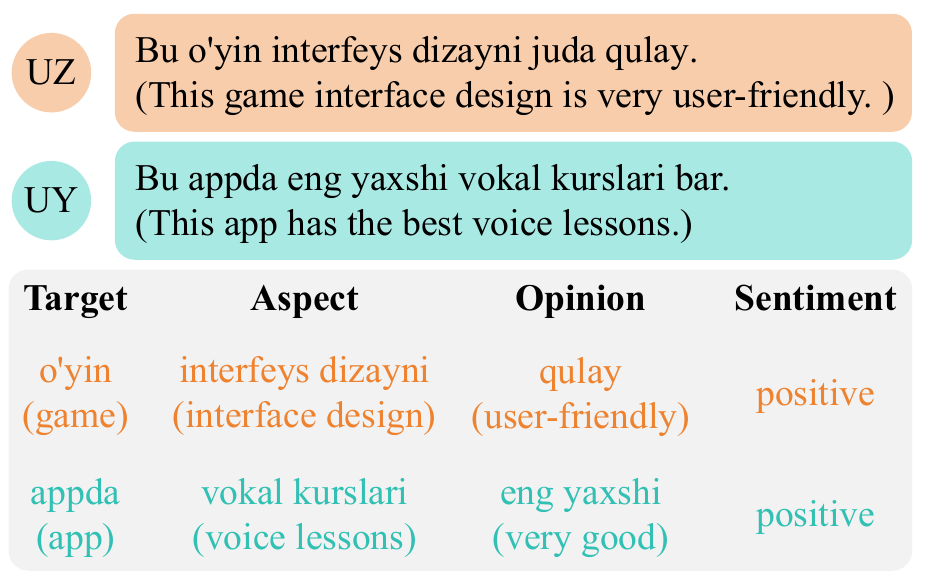}
\caption{Input and output examples of our dataset. The text in parentheses represents the corresponding English translation.}
\label{Fig1}
\end{figure}

To address these two shortcomings, we constructe \textbf{L}ow-resource languages \textbf{A}spect-based \textbf{S}entiment \textbf{Q}uadruple (LASQ), a new benchmark dataset specifically designed for aspect-based sentiment quadruple detection in under-represented languages. As shown in Figure~\ref{Fig1}, LASQ covers two typologically distinct languages, Uzbek (UZ) and Uyghur (UY), and emphasizes comprehensive structured annotation, including \emph{target}, \emph{aspect}, \emph{opinion}, and \emph{sentiment}.

LASQ consists of naturally occurring user reviews independently collected from Google Play Store applications, forming a bilingual corpus in Uzbek and Uyghur. Due to the extremely limited availability of Uyghur-script user reviews in practice, the Uyghur subset was gathered via targeted crawling and filtering to ensure sufficient quality and diversity for benchmarking. After careful manual cleaning and sentence segmentation, the LASQ corpus includes two subsets: Uzbek and Uyghur. The Uzbek subset contains 3,064 documents with 11,728 sentiment quadruples, while the Uyghur subset contains 3,028 documents with 11,543 sentiment quadruples.To ensure annotation reliability, each sentence was annotated by multiple annotators and further reviewed by experts, achieving a Cohen's kappa of 0.85, which indicates strong agreement. Overall, LASQ differs from prior datasets by jointly supporting target-aware quadruple annotation and bilingual low-resource coverage under a unified annotation schema, thereby providing a richer and more practical resource for advancing structured sentiment extraction in minor languages.

Next, we constructed a benchmark for our LASQ dataset. In the high-resource ABSA field, researchers have proposed various methods: 1) Span-based methods \citep{www_baseline,ACOS,Span-ASTE} extract spans from the text and identify relations between spans to construct quadruples. This method suffers from error propagation and is time-consuming due to the need to enumerate spans. 2) Generative methods \citep{ParaPhrase,jiegoushengcheng,shengchengshi} directly utilize pre-trained generative models or large models to generate quadruples. This method is effective in high-resource languages, but suffers from severe illusion problems for low-resource languages like Uzbek. 3) Grid-tagging-based methods \citep{h2dt,meatwp} are favored by researchers. This method converts sentences into two-dimensional tables, thereby simultaneously extracting spans and relations, avoiding error propagation, and eliminating the need for span enumeration.

Therefore, our benchmark adopts the grid-tagging-based method and has been improved based on the characteristics of low-resource languages. Considering that both Uzbek and Uyghur belong to the Altaic language family and have typical cohesive features: a word stem can be followed by a sequence of derivational and inflectional suffixes to express grammatical information such as case, number, tense, and person. This results in high morphological complexity and severe vocabulary sparsity, where the same lemma may appear in many surface forms across contexts, substantially increasing OOV rates and modeling difficulty, especially when annotated data are limited \citep{uy1,Turkish,nianzhuoyu}. Therefore, our key motivation is to introduce stronger structural inductive bias for low-resource Altaic languages. After text encoding, we incorporate a \textbf{S}yntax \textbf{K}nowledge \textbf{E}mbedding \textbf{M}odule (SKEM) that injects part-of-speech (POS) knowledge and dependency knowledge, alleviating morphology-induced sparsity and enabling the model to more stably capture cross-token semantic and sentiment associations. Specifically, we first encode the input sequence into contextual representations, then use SKEM to map POS tags and dependency structures into learnable structured embeddings and fuse them with semantic representations through interaction, and finally perform unified structured prediction for target--aspect--opinion--sentiment quadruples within a single framework.

We compared our approach with eight baseline models, including two large language models (LLMs), and found our method to be the best in both languages. In Uzbek and Uyghur, our method improved the F1 score by 1.28\% and 1.86\% respectively compared to the best baseline models. We further validated the effectiveness of our proposed module using various analytical methods, including ablation experiments.

In summary, our contributions are threefold:
\begin{itemize}
\item We constructed the first \textbf{L}ow-resource languages \textbf{A}spect-based \textbf{S}entiment \textbf{Q}uadruple dataset, named LASQ, which include two low-resource languages: Uzbek and Uyghur, and provides fine-grained target-aspect-opinion-sentiment quadruple annotations.

\item We designe a syntactic knowledge-enriched grid-tagging model by incorporating part-of-speech (POS) and dependency knowledge via a \textbf{S}yntax \textbf{K}nowledge \textbf{E}mbedding \textbf{M}odule (SKEM), introducing effective structural inductive bias to mitigate vocabulary sparsity and improve cross-token sentiment association modeling.

\item Our method outperforms six competitive baseline models and two state-of-the-art LLMs in both Uzbek and Uyghur languages. Furthermore, we conducted additional analytical experiments to validate the effectiveness of the proposed modules.
\end{itemize}

\section{Related work}\label{sec2}

\subsection{ABSA for High-resource Languages}
Sentiment analysis has long been a fundamental task in natural language processing~\citep{RW1,RW2}, with broad applications in sentiment-aware chatbots~\citep{RW3,RW4}, recommender systems~\citep{rw5,rw6}, and question answering~\citep{RW7,RW8}. As a fine-grained branch of sentiment analysis, aspect-based sentiment analysis (ABSA) aims to identify aspect-related sentiment and has evolved from early polarity classification toward structured prediction involving multiple sentiment components~\citep{RW9,RW10,RW11}. For high-resource languages such as English and Chinese, abundant annotated data and relatively mature toolchains (e.g., tokenization, parsing, and lexicons) have continuously driven progress.

Early ABSA research mainly relied on sentiment lexicons, rule-based patterns, and classical machine learning models such as Naive Bayes and SVM, where handcrafted lexical/syntactic features played a central role~\citep{sa2,sa3,EN2}. Subsequently, deep learning models (e.g., CNN/LSTM with attention) became dominant by learning contextual representations and modeling aspect--sentiment interactions more effectively~\citep{cnn,lstm,ZH2,ZH1}. With the advent of pre-trained language models (e.g., BERT, RoBERTa, ERNIE), ABSA performance in high-resource languages further improved due to stronger contextual modeling and transferability~\citep{Bert,roberta,ernie,ZH3,ZH4,liaoRoBerta}. However, these sequence models all suffer from difficulty in capturing complex syntactic interactions. Therefore, some studies have mitigated this problem by incorporating syntactic knowledge to explicitly model long-distance syntactic dependencies between target, aspect, and opinion \citep{syntactic_1,syntactic_2,syntactic_3}. More recently, generative pre-trained models (e.g., T5/BART) have also been explored by reformulating ABSA and related extraction tasks as text generation, enabling unified modeling via prompts or templates~\citep{EN5,EN4}.

In parallel, instruction-following large language models (LLMs) such as GPT-series and DeepSeek have shown promising zero-/few-shot capabilities for ABSA in high-resource languages through prompting and in-context learning~\citep{LLMABSA1,LLMABSA2}. Nevertheless, their advantages do not always transfer to low-resource languages or complex structured ABSA settings, where limited linguistic coverage, weak task grounding, and instability in structured outputs can lead to degraded extraction quality~\citep{LLMABSA3,LLMABSA4,LLMABSA5}. These observations motivate our focus on low-resource ABSA quadruple extraction and the need for dedicated benchmarks and structure-aware modeling.

\subsection{ABSA for Low-resource Languages} 
Despite rapid progress in high-resource languages, ABSA for low-resource languages remains constrained by scarce annotations, limited linguistic tools, and domain mismatch. Therefore, a major research line focuses on \emph{resource construction} and \emph{model adaptation} under limited supervision.

On the resource side, several datasets and benchmarks have been created to enable aspect-level sentiment modeling in specific low-resource languages. For Vietnamese, \citet{Vietnamese} constructed a span-level ABSA dataset that annotates opinion spans together with aspect--sentiment labels, and provided sequence-tagging baselines for Vietnamese reviews; however, it is not designed for explicit target--aspect--opinion--polarity quadruple extraction. For Telugu, \citet{Telugu} built an ABSA resource focusing on aspect term extraction and polarity classification, which supports aspect-level sentiment prediction but does not include opinion span extraction. For Urdu, \citet{Urdu} introduced an ABSA dataset with aspect-related components and evaluated classical as well as neural classifiers on Urdu social media text, yet the annotations remain centered on aspect-level polarity rather than end-to-end extraction of opinion expressions. For Indonesian, \citet{Indonesian} adapted a generative multitask prompting framework and reported results on unified ABSA and triplet-style extraction settings, which still differs from our setting by not requiring full quadruple-level supervision with explicit target and opinion spans.

On the modeling side, low-resource ABSA commonly relies on transfer learning and structure-aware inductive biases to reduce data demands \citep{xiangguan2,xiangguan3,xiangguan4}. 
Multilingual pre-trained language models and cross-lingual transfer are frequently adopted to improve generalization across domains and languages \citep{xiangguan4,xiangguan2}. 
In addition, syntactic priors such as dependency-based graph modeling can help stabilize aspect--opinion interactions when supervision is limited \citep{xiangguan,xiangguan1}, 
while end-to-end structured prediction paradigms aim to jointly model multiple sentiment components to mitigate pipeline error propagation \citep{xiangguan5,xiangguan6}. 
These trends highlight the continued need for dedicated low-resource benchmarks and structure-aware approaches for complex ABSA extraction tasks.

\begin{figure}[t]
\centering
\includegraphics[width=0.8\textwidth]{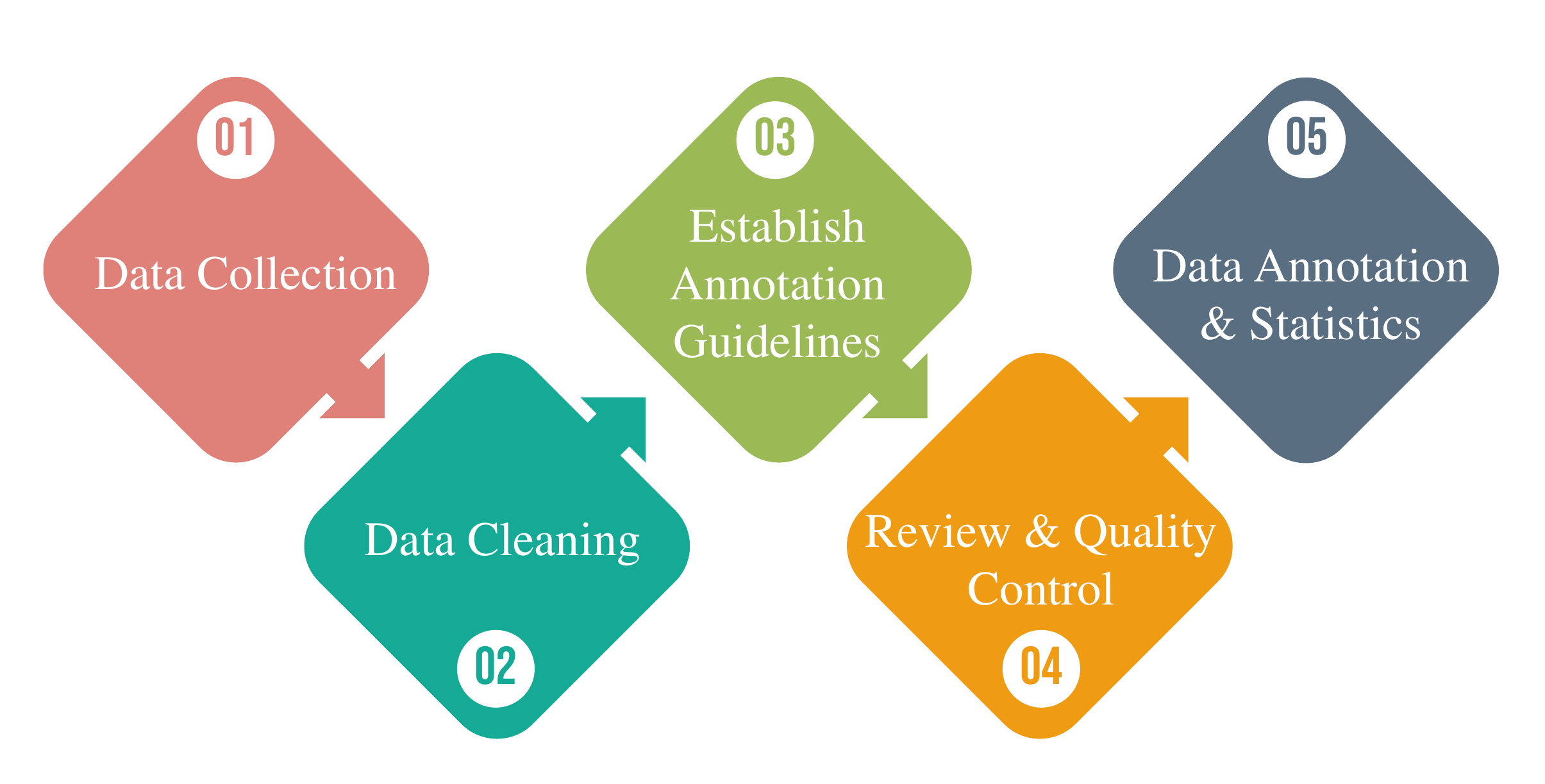}
\caption{LASQ dataset construction flowchart.}
\label{anno}
\end{figure}

\section{ Dataset construction}\label{sec:3}

Our dataset construction process is shown in Figure ~\ref{anno}, which includes five steps: data collection, data cleaning, establish annotation guidelines, review and quality control, and data annotation and statistics. We will now describe each step in detail.

\subsection{Data Collection}
The dataset construction process commenced with the collection of 50,000 user reviews sourced from software and game applications on the Google Play Store platform\footnote{\url{https://play.google.com}}. 
To support bilingual ABSA research, we independently crawled reviews written in Uzbek and Uyghur, targeting applications frequently used by Uzbek- and Uyghur-speaking users. 
This design ensures that both language subsets are naturally occurring user-generated content and representative of their respective linguistic and cultural contexts.

To ensure diversity and representativeness, reviews were drawn from a wide range of application categories, including productivity tools, entertainment platforms, and utility software. 
The selection emphasized collecting a balanced corpus that captures varying user experiences, sentiments, and app functionalities, covering aspects such as performance, usability, and customer support.

The collected reviews served as the foundation for subsequent processing and refinement stages, aimed at creating a high-quality bilingual dataset tailored to fine-grained aspect-based sentiment analysis.

\subsection{Data Cleaning}
In the data cleaning phase, we systematically processed the raw reviews in both languages to enhance data quality and consistency. 
This procedure removed duplicated entries, advertisements, texts containing undesirable characters, and incomplete or overly short sentences. 
We further filtered out sentences lacking explicit sentiment-bearing components, retaining only samples with identifiable targets/aspects/opinions suitable for ABSA annotation.

After cleaning and filtering, we obtained a refined corpus consisting of Uzbek and Uyghur app reviews. 
Each review was curated to meet the requirements of fine-grained ABSA, providing a robust foundation for subsequent annotation.

\subsection{Establish Annotation Guidelines}
To ensure consistent and reproducible sentiment quadruple annotation on app reviews, we established unified annotation guidelines before large-scale labeling. As illustrated in Figure \ref{Fig1}, each sentence is annotated with a quadruple consisting of Target, Aspect, Opinion, and Sentiment, where Target denotes the entity being evaluated, Aspect refers to the specific attribute or feature of the target, Opinion is the explicit sentiment-bearing expression describing the aspect, and Sentiment is labeled as positive or negative. The guidelines were designed to be consistent across two typologically different languages, Uzbek and Uyghur, so that the same labeling principles apply despite linguistic variation.
We first annotate Target, Aspect, and Opinion, and then annotate Sentiment based on the sentiment expressed by the Opinion. Specifically, for Target, we only annotate the specific entity, the application name or game name (including its full name, abbreviation, nickname, etc.), and do not annotate pronouns. For Aspect, we only annotate aspects of the already annotated targets, and aspects can be missing. For Opinion, we only annotate opinions of the targets or aspects that have been annotated.

\subsection{Review and Quality Control}
We implemented a multi-tier quality control mechanism to resolve boundary discrepancies and maintain annotation consistency. 
For each instance, disagreements among annotators were first discussed within the annotation group; unresolved cases were escalated to domain experts for final adjudication. 
Meanwhile, the annotation guideline was iteratively refined based on recurring disagreement patterns to improve consistency. To quantify inter-annotator reliability, Cohen's kappa coefficient ($\kappa$) was computed at the quadruple level:
$$\kappa = \frac{P_o - P_e}{1 - P_e},$$
where $P_o$ denotes the observed agreement and $P_e$ denotes the chance agreement. 
The observed agreement rates were 90\% for targets, 88\% for aspects, 87\% for opinions, and 85\% for polarities, yielding an overall $\kappa$ score of 85\%, indicating substantial inter-annotator consistency. 
Remaining discrepancies were resolved through expert review and consensus.

\subsection{Data Annotation and Statistics}
We used the BRAT annotation tool\footnote{\url{https://brat.nlplab.org/index.html}} for annotate targets, aspects, opinions, and sentiment.Specifically, our annotation process consists of two steps:

1) Entity Annotation: We define three entity tags: \texttt{TAR}, \texttt{ASP}, and \texttt{OPIN}, to annotate these entities.

2) Relation Annotation: After entity annotation, we need to annotate sentiment and construct quadruples. Therefore, we define four relation tags: \texttt{TAR-ASP}, \texttt{ASP-OPIN}, \texttt{POS}, and \texttt{NEG}.

\begin{figure}[t]
\centering
\includegraphics[width=1\textwidth]{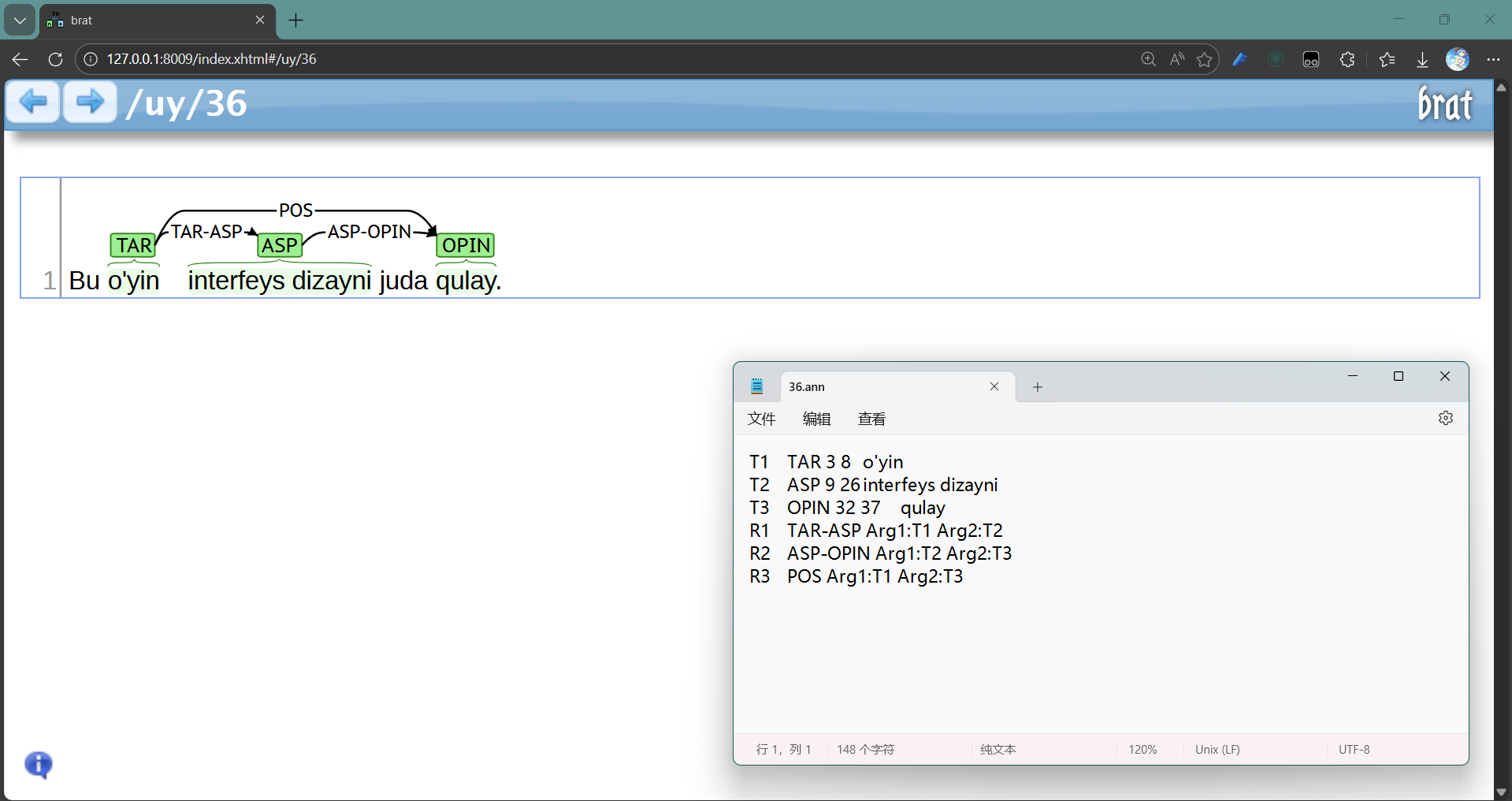}
\caption{An example of annotation on the brat platform and its saved ANN format annotation result (bottom right corner of the figure). The example in the image is in Uzbek and means "This game interface design is very user-friendly."}
\label{brat}
\end{figure}

Figure \ref{brat} shows an example of annotation on the brat platform and its saved ANN format annotation result, "o'yin (game)" is annotated as \texttt{TAR}, "interfeys dizayni (interface design)" is annotated as \texttt{ASP}, and "qulay (user-friendly)" is annotated as \texttt{OPIN}. Then, \texttt{TAR-ASP} relation arc is annotated between "o'yin" and "interfeys dizayni", \texttt{ASP-OPIN} relation arc between "interfeys dizayni" and "qulay", and \texttt{POS} relation arc between "o'yin" and "qulay". After the annotation is completed, an annotation result file in ANN format will be generated, which will then be converted to JSON format through post-processing. Finally, we randomly divided the labeled dataset into training, validation, and test sets in an 8:1:1 ratio. Specific dataset statistics are presented in the table \ref{tab:statistics}. The Uzbek and Uyghur datasets have a total sample size of 3,064 and 3,028 respectively, containing 11,728 (7,796 positive and 3,688 negative) and 11,543 (7,763 positive and 3,680 negative) quadruplets.

\begin{table*}[t]
\fontsize{12}{16}\selectfont
\setlength{\tabcolsep}{0.8mm}
\setlength{\arrayrulewidth}{0.1mm}
\centering
\resizebox{\textwidth}{!}{
\begin{tabular}{cccccccccccccccccccccc} 
\hline
& & \phantom{}&\multirow{2}{*}{\textbf{\# Doc.}}&\phantom{}&\multirow{2}{*}{\textbf{\# Sent.}}&\phantom{}&\multirow{2}{*}{\textbf{\# Token}} & \phantom{}& \multicolumn{5}{c}{\bf \# Entity} & \phantom{}&\multicolumn{5}{c}{\bf \# Relation} & \phantom{}&\multirow{2}{*}{\# \bf Quad.} \\
\cmidrule(lr){10-14}\cmidrule(lr){16-20}
& & \phantom{}&&\phantom{}& &\phantom{}&&\phantom{}& \bf T& \phantom{}&\bf A& \phantom{}& 
 \bf O & \phantom{}& \bf TA& \phantom{}&\bf TO & \phantom{}& \bf AO & \phantom{}& \\
\hline
 \multirow{4}{*}{\bf UZ} & \bf Train&\phantom{} &2,451&\phantom{}&6,146&\phantom{}& 101,061&\phantom{} & 2,471 & \phantom{}& 7,075 & \phantom{}& 8,697&\phantom{}&7,884& \phantom{}&8,712& \phantom{}&9,348&\phantom{}&9,384 \\
 & \bf Dev&\phantom{} &306&\phantom{}&780&\phantom{} &12,705&\phantom{} & 307 & \phantom{}& 841 & \phantom{}& 1,031&\phantom{}&943& \phantom{}&1,031& \phantom{}&1,097&\phantom{}&1,100 \\
 & \bf Test&\phantom{} & 307&\phantom{}&744&\phantom{}&12,561&\phantom{} & 307 & \phantom{}& 955 & \phantom{}& 1,130&\phantom{}&1,058& \phantom{}&1,130& \phantom{}&1,239&\phantom{}&1,244 \\
 \cdashline{2-22}
 & \bf Total&\phantom{} & 3,064&\phantom{}&7,670&\phantom{}&126,327&\phantom{} & 3,085 & \phantom{}& 8,871 & \phantom{}& 10,858&\phantom{}&9,885& \phantom{}&10,873& \phantom{}&11,684&\phantom{}&11,728 \\
 \hline
 \multirow{4}{*}{\bf UY} & \bf Train&\phantom{}&2,422 &\phantom{}&8,539&\phantom{}& 90,184&\phantom{} & 2,427 & \phantom{}& 6,767 & \phantom{}& 8,248&\phantom{}&7,577& \phantom{}&8,250& \phantom{}&9,202&\phantom{}&9,231 \\
 & \bf Dev&\phantom{} & 302&\phantom{}&1,017&\phantom{}&11,343&\phantom{} & 302 & \phantom{}& 845 & \phantom{}& 1,066&\phantom{}&954& \phantom{}&1,066& \phantom{}&1,159&\phantom{}&1,161 \\
 & \bf Test&\phantom{} &304&\phantom{}&1,063&\phantom{}& 11,203&\phantom{} & 304 & \phantom{}& 847 & \phantom{}& 1,025&\phantom{}&950& \phantom{}&1,025& \phantom{}&1,149&\phantom{}&1,151 \\
\cdashline{2-22}
 & \bf Total&\phantom{} & 3,028&\phantom{}&10,619&\phantom{}&112,730&\phantom{} & 3,033 & \phantom{}& 8,459 & \phantom{}& 10,339&\phantom{}&9,481& \phantom{}&10,341& \phantom{}&11,510&\phantom{}&11,543 \\
 \hline
\end{tabular}
}
\caption{Statistics of our dataset. T/A/O represents target, aspect and opinion, and TA/TO/AO represents target-aspect pair, target-opinion pair and aspect-opinion pair. (Doc.: Document, Sent.: Sentence, Quad.: Quadruple).
}
\label{tab:statistics}
\end{table*}

\section{Methodology}\label{sec:4}

\subsection{Task Definition}
Our task goal is to extract the aspect-level sentiment quadruple set $Q=\{q_{1},q_{2},\dots,q_{n_{q}}\}\in\mathbb{R}^{n_{q}}$ from the given text $X=\{x_{1},x_{2},\dots,x_{n_{x}}\}\in\mathbb{R}^{n_{x}}$. For the $i$-th quadruple $q_{i}$ includes the target $t_{i}=\{b_{i}^{t},e_{i}^{t}\}$, aspect $a_{i}=\{b_{i}^{a},e_{i}^{a}\}$, opinion $o_{i}=\{b_{i}^{o},e_{i}^{o}\}$ and sentiment $s_{i}\in\{pos, neg\}$, where $b_{i}^{*}$ and $e_{i}^{*}$ represent the start and end indexes,  $*=\{t,a,o\}$, $pos$ and $neg$ represent positive and negative sentiments respectively.

\begin{figure*}[t]
\centering
\includegraphics[width=1\textwidth]{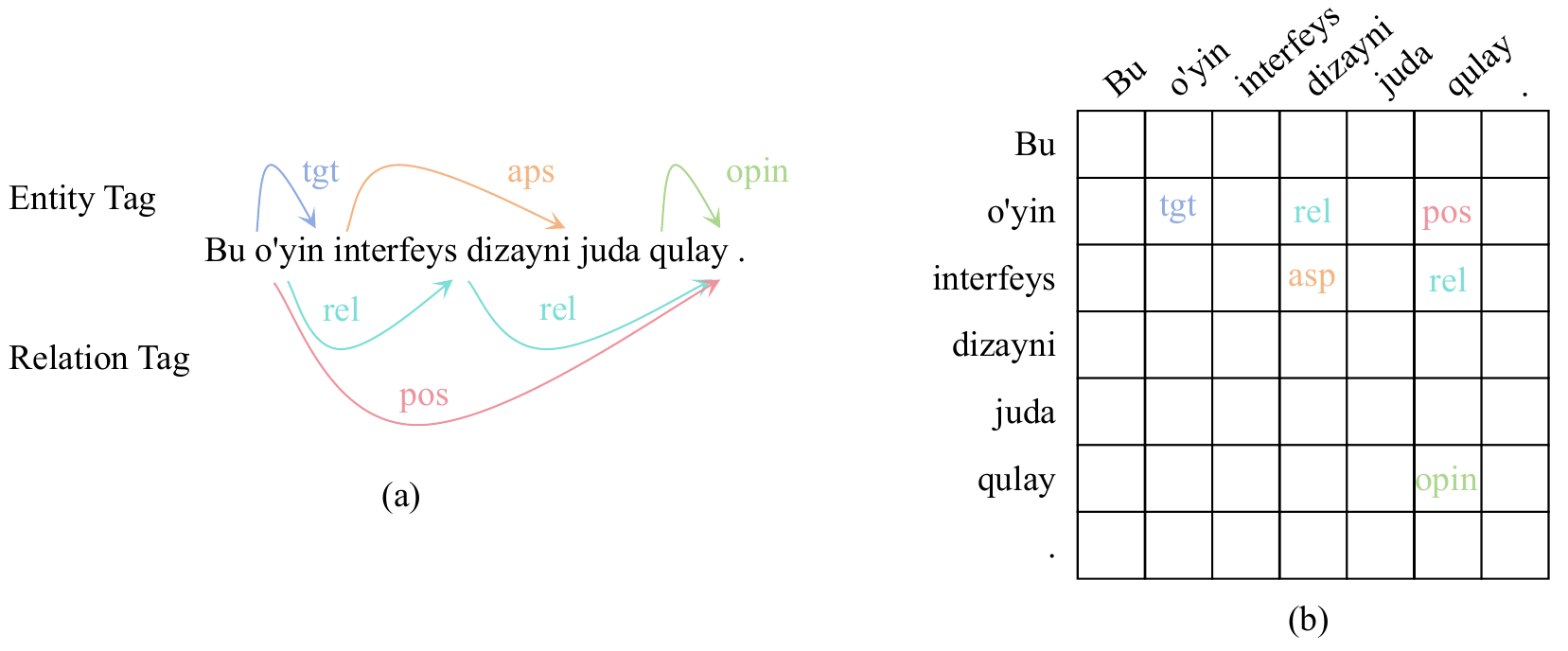}
\caption{(a) Our grid-tagging system. (b) Grid-tagging system in grid. The example in the image is in Uzbek and means "This game interface design is very user-friendly."} 
\label{grid}
\end{figure*}

\subsection{Tagging System}
We transform quadruple extraction into a grid-tagging relation extraction task. 
Specifically, we construct two grid-tagging systems: one is an entity grid for extracting targets, aspects and opinions, and the other is a relation grid for extracting relations between them and performing sentiment classification. Finally, the final quadruple results are obtained by further decoding the grid.

For the entity grid, we define four tags: 
\begin{itemize}
    \item \texttt{tgt}: It is used to identify the target entity, with the row and column where the tag is located representing the head and tail of the entity, respectively.
    \item \texttt{asp}: It is used to identify the aspect entity, with the tag in the grid having the same meaning as \texttt{tgt}.
    \item \texttt{opin}: It is used to identify the opinion entity, with the tag in the grid having the same meaning as \texttt{tgt}.
    \item \texttt{none}: Except for the above tags, the rest of the positions in the grid are all this tag.
\end{itemize}
For the relation grid, we define five tags:
\begin{itemize}
    \item \texttt{rel}: It indicates that when two entities (such as target and aspect, or aspect and opinion) have this relation, they belong to the same quadruple. The row and column where the tag is located represent the heads of the two entities, respectively.
    \item \texttt{pos}: It indicates that the sentiment category of the quadruple is positive, with the row and column where the tag is located representing the heads of the target and opinion, respectively.
    \item \texttt{neg}: It indicates that the sentiment category of the quadruple is negative, with the tag in the grid having the same meaning as \texttt{pos}.
    \item \texttt{none}: Except for the above tags, the rest of the positions in the grid are all this tag.
\end{itemize}

As shown in Figure \ref{grid}, we first decode the target "o'yin (game)" through the \texttt{tgt} tag in the entity grid. Similarly, we decode the aspect "interfeys dizayni (interface design)" and the opinion "qulay (user-friendly)" through the \texttt{asp} tag and \texttt{opi} tag. Then we decode the triples (o'yin, interfeys dizayni, qulay) through the \texttt{rel} tag in the relation grid. Finally, we decode the quadruplets (o'yin, interfeys dizayni, qulay, positive) through the \texttt{pos} tag.

\begin{figure*}[t]
\centering
\includegraphics[width=1\textwidth]{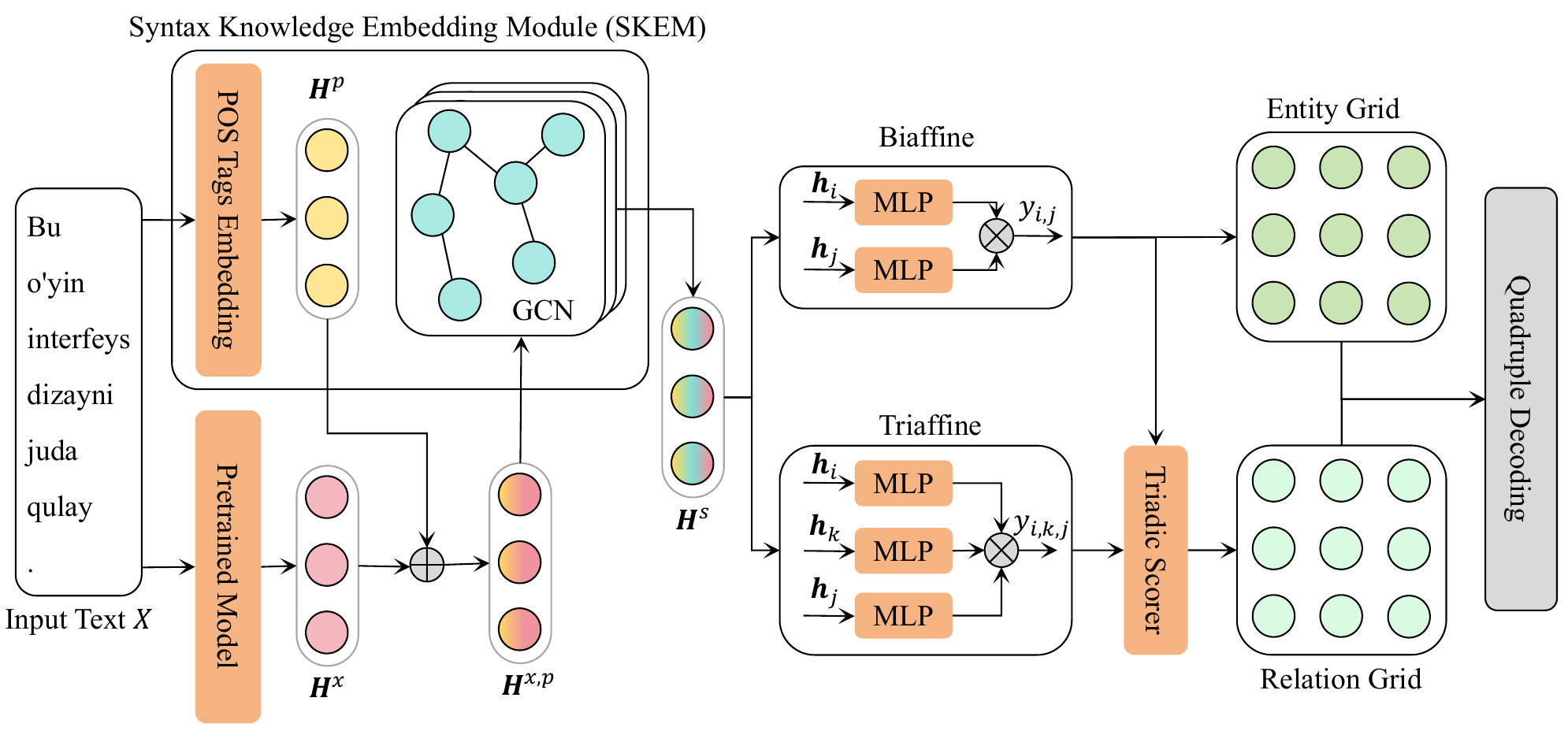}
\caption{The overall architecture of our model. $ \bigoplus $ and $\bigotimes$ represent matrix concatenation and matrix multiplication. The example in the image is in Uzbek and means "This game interface design is very user-friendly."
} 
\label{model_frame}
\end{figure*}

\subsection{Model Architecture}
Our model architecture is shown in Figure \ref{model_frame}, which includes four modules: text encoder, syntax knowledge embedding module , entity extraction, and relation extraction.

\subsubsection{Text Encoder}
We use a pre-trained model to encode the text, such as RoBERTa  \citep{roberta}, and get the text embedding:
\begin{equation}
\label{deqn_ex9a}
\begin{gathered}
\boldsymbol{H}^{x}=\{\boldsymbol{h}_{1}^{x},\boldsymbol{h}_{2}^{x},\dots,\boldsymbol{h}_{n_{x}}^{x}\}\in\mathbb{R}^{n_{x}\times h_{x}}, \\
\end{gathered}
\end{equation}
Where $\boldsymbol{h}_{i}^{x}$ represents the embedding of the $i$-th word and $h_{x}$ represents the dimension of the word embedding.

\subsubsection{Syntax Knowledge Embedding Module (SKEM)}
Since our dataset is multilingual and the underlying syntactic knowledge of different languages is similar, we designed a SKEM to incorporate rich syntactic knowledge into the model. First, for the POS knowledge $P=\{p_{1},p_{2},\dots,p_{n_{x}}\}\in\mathbb{R}^{n_{x}}$, we get the POS embedding through the POS embedding layer:
\begin{equation}
\label{deqn_ex9b}
\begin{gathered}
\boldsymbol{H}^{p}=\{\boldsymbol{h}_{1}^{p},\boldsymbol{h}_{2}^{p},\dots,\boldsymbol{h}_{n_{x}}^{p}\}\in\mathbb{R}^{n_{x}\times h_{p}}, \\
\end{gathered}
\end{equation}
Where $\boldsymbol{h}_{i}^{p}$ represents the POS embedding corresponding to the $i$-th word, and $h_{p}$ represents the dimension of the POS embedding. Then we concatenate $\boldsymbol{H}^{x}$ and $\boldsymbol{H}^{p}$ to get the word embedding $\boldsymbol{H}^{x,p}$ that integrates POS knowledge:  
\begin{equation}
\label{deqn_ex9a}
\begin{gathered}
\boldsymbol{H}^{x,p}=\{\boldsymbol{H}^{x};\boldsymbol{H}^{p}\}\in\mathbb{R}^{n_{x}\times (h_{x}+h_{p})}, \\
\end{gathered}
\end{equation}
Where $\{\cdot;\cdot\}$ represents the concatenation operation. Then we use graph convolution network (GCN)  \citep{gcn} to integrate dependency knowledge into $\boldsymbol{H}^{x,p}$ to obtain the word embedding $\boldsymbol{H}^{s}\in\mathbb{R}^{n_{x}\times h_{s}}$ with rich syntactic knowledge, as follows:
\begin{equation}
\label{deqn_ex4a}
\begin{gathered}
\boldsymbol{H}^{s}=\sigma(A\boldsymbol{H}^{x,p}\boldsymbol{W}_{GCN})=\{\boldsymbol{h}_{1}^{s},\boldsymbol{h}_{2}^{s},\dots,\boldsymbol{h}_{n_{x}}^{s}\}
\end{gathered}
\end{equation}
Where $\sigma$ is the activation function and $\boldsymbol{W}_{GCN}\in\mathbb{R}^{(h_{x}+h_{p})\times h_{s}}$ is the trainable weight. $A\in\mathbb{R}^{n_{x}\times n_{x}}$ is the adjacency matrix containing the dependencies, and its element $a_{i,j}$ in the $i$-th row and $j$-th column is:
\begin{equation}
\label{deqn_ex4a}
\begin{gathered}
\begin{cases}a_{i,j}=a_{j,i}=1, x_{i}\text{ and } x_{j} \text{ have relations},\\
a_{i,j}=a_{j,i}=0, x_{i}\text{ and }x_{j}\text{ don't have relations}.
\end{cases}
\end{gathered}
\end{equation}

\subsubsection{Entity Extraction}
In the entity extraction stage, we first map $\boldsymbol{H}^{s}$ to the entity space through a linear layer to obtain $\boldsymbol{H}^{e}\in\mathbb{R}^{n_{x}\times h_{e}}$:
\begin{equation}
\label{deqn_ex4a}
\begin{gathered}
\boldsymbol{H}^{e}=\boldsymbol{H}^{s}\boldsymbol{W}^{e}+\boldsymbol{b}^{e}=\{\boldsymbol{h}_{1}^{e},\boldsymbol{h}_{2}^{e},\dots,\boldsymbol{h}_{n_{x}}^{e}\},
\end{gathered}
\end{equation}
where $\boldsymbol{W}^{e}\in\mathbb{R}^{h_{s}\times h_{e}}$ and $\boldsymbol{b}^{e}\in\mathbb{R}^{ h_{e}}$ are trainable parameters. Then we use biaffine \citep{Biaffine} to obtain the score of the word pair $(\boldsymbol{h}_{i}^{e}, \boldsymbol{h}_{j}^{e})$ in the entity grid according to the label level:
\begin{equation}
\label{biaffine_eq}
\begin{gathered}
\hat{\boldsymbol{h}}_{i}^{l}=\text{MLP}_{1}^{l}(\boldsymbol{h}_{i}^{e}),\\
\hat{\boldsymbol{h}}_{j}^{l}=\text{MLP}_{2}^{l}(\boldsymbol{h}_{j}^{e}),\\
y^{l}_{i,j}=(\hat{\boldsymbol{h}}_{i}^{l})^{\top}\boldsymbol{W}^{l}\hat{\boldsymbol{h}}_{j}^{l},\\
\end{gathered}
\end{equation}
where $\boldsymbol{W}^{l}\in\mathbb{R}^{h_{e}\times h_{e}}$ is trainable parameters, and $l\in\{\text{\texttt{tgt}},\text{\texttt{asp}},\text{\texttt{opin}},\text{\texttt{none}}\}$. Finally, we can get the probability distribution $p^{e}_{i,j}$ of the label for each word pair $(\boldsymbol{h}_{i}^{e}, \boldsymbol{h}_{j}^{e})$ in the entity matrix:
\begin{equation}
\label{deqn_ex4a}
\begin{gathered}
\boldsymbol{p}^{e}_{i,j}=\text{Softmax}([y_{i,j}^{\text{\texttt{tgt}}};y_{i,j}^{\text{\texttt{asp}}};y_{i,j}^{\text{\texttt{opin}}};y_{i,j}^{\text{\texttt{none}}}]).\\
\end{gathered}
\end{equation}

\subsubsection{Relation Extraction}
Similarly, in the entity extraction stage, we first map $\boldsymbol{H}^{s}$ to the relation space through a linear layer to obtain $\boldsymbol{H}^{r}\in\mathbb{R}^{n_{x}\times h_{r}}$:
\begin{equation}
\label{deqn_ex4a}
\begin{gathered}
\boldsymbol{H}^{r}=\boldsymbol{H}^{s}\boldsymbol{W}^{r}+\boldsymbol{b}^{r}=\{\boldsymbol{h}_{1}^{r},\boldsymbol{h}_{2}^{r},\dots,\boldsymbol{h}_{n_{x}}^{r}\},
\end{gathered}
\end{equation}
where $\boldsymbol{W}^{r}\in\mathbb{R}^{h_{s}\times h_{r}}$ and $\boldsymbol{b}^{r}\in\mathbb{R}^{h_{r}}$ are trainable parameters. Then we obtain the score $\hat{y}^{q}_{i,j}$ of each word pair $(\boldsymbol{h}_{i}^{r}, \boldsymbol{h}_{j}^{r})$ in the relation grid:
\begin{equation}
\label{deqn_ex4a}
\begin{gathered}
\hat{\boldsymbol{h}}_{i}^{q}=\text{MLP}_{1}^{q}(\boldsymbol{h}_{i}^{r}),\\
\hat{\boldsymbol{h}}_{j}^{q}=\text{MLP}_{2}^{q}(\boldsymbol{h}_{j}^{r}),\\
\hat{y}^{q}_{i,j}=(\hat{\boldsymbol{h}}_{i}^{q})^{\top}\boldsymbol{W}^{q}\hat{\boldsymbol{h}}_{j}^{q},\\
\end{gathered}
\end{equation}
where $q\in\{\text{\texttt{rel}},\text{\texttt{pos}},\text{\texttt{neg}},\text{\texttt{none}}\}$. Next we use triaffine \citep{triaffine} at the label level to determine the score of each triple $(\boldsymbol{h}_{i}^{r}, \boldsymbol{h}_{k}^{r}, \boldsymbol{h}_{j}^{r})$:
\begin{equation}
\label{deqn_ex4a}
\begin{gathered}
\tilde{\boldsymbol{h}}^{q}_{i},\tilde{\boldsymbol{h}}^{q}_{k},\tilde{\boldsymbol{h}}^{q}_{j}=\text{MLP}_{3}^{q}(\boldsymbol{h}^{r}_{i}),\text{MLP}_{4}^{q}(\boldsymbol{h}^{r}_{k}),\text{MLP}_{5}^{q}(\boldsymbol{h}^{r}_{j}),\\
\tilde{y}^{q}_{i,k,j}=\boldsymbol{W}^{r}
\begin{bmatrix}
  \tilde{\boldsymbol{h}}^{q}_{i} \\
  1
\end{bmatrix}
\tilde{\boldsymbol{h}}^{q}_{k}\tilde{\boldsymbol{h}}^{q}_{j},\\
y^{q}_{i,k,j}=\frac{\text{exp}(\tilde{y}^{q}_{i,k,j})}{\sum_{q}\text{exp}(\tilde{y}^{q}_{i,k,j})},
\end{gathered}
\end{equation}
where $\boldsymbol{W}^{r}\in\mathbb{R}^{h_{r}\times h_{r}\times (h_{r}+1)}$ is trainable parameters. We then use the triaffine score as a bridge to calculate the final word pair $(\boldsymbol{h}_{i}^{r}, \boldsymbol{h}_{j}^{r})$ score:
\begin{equation}
\label{deqn_ex4a}
\begin{gathered}
y_{i,j}^{q}=\hat{y}_{i,j}^{q}+\sum_{k}(\hat{y}_{i,k}^{q}+\hat{y}_{k,j}^{q})*y^{q}_{i,k,j}
\end{gathered}
\end{equation}
Finally, we can get the probability distribution $p^{r}_{i,j}$ of the label for each word pair $(\boldsymbol{h}_{i}^{r}, \boldsymbol{h}_{j}^{r})$ in the entity matrix:
\begin{equation}
\label{deqn_ex4a}
\begin{gathered}
\boldsymbol{p}^{r}_{i,j}=\text{Softmax}([y_{i,j}^{\text{\texttt{rel}}};y_{i,j}^{\text{\texttt{pos}}};y_{i,j}^{\text{\texttt{neg}}};y_{i,j}^{\text{\texttt{none}}}]).\\
\end{gathered}
\end{equation}

\subsection{Learning}
We jointly optimize our model by entity and relation grid as follows::
\begin{equation}
\label{deqn_ex4a}
\begin{gathered}
\mathcal{L}^{e}=-\frac{1}{n^{2}}\sum_{i=1}^{n}\sum_{j=1}^{n}{g}^{e}_{i,j}\text{log}\boldsymbol{p}^{e}_{i,j},\\
\mathcal{L}^{r}=-\frac{1}{n^{2}}\sum_{i=1}^{n}\sum_{j=1}^{n}{g}^{r}_{i,j}\text{log}\boldsymbol{p}^{r}_{i,j},\\
\mathcal{L}=\mathcal{L}^{e}+\mathcal{L}^{r}.\\
\end{gathered}
\end{equation}
Where $g^{e}_{i,j}$ and $g^{r}_{i,j}$ represent the golden labels of entities and relations, $\mathcal{L}^{e}$ and $\mathcal{L}^{r}$ represent the losses of entities and relations, and $\mathcal{L}$ represents the total loss.

\section{Experimental Settings}\label{sec:5}
\subsection{Baselines}
We compared eight baseline models, including two state-of-the-art LLMs.
\textbf{EC-ACOS} \citep{ACOS} is a span-based method that first extracts aspect-opinion pairs and then predicts the category and sentiment of these span pairs.
\textbf{Span-ASTE} \citep{Span-ASTE} also a span-based method that explicitly considers the interactions between spans to ensure better emotional consistency.
\textbf{ParaPhrase} \citep{ParaPhrase} is a generative method that transforms the ABSA task into a paraphrase generation process, directly generating quadruples in an end-to-end manner.
\textbf{Meta-WP} \citep{meatwp} is a grid-tagging-based method that achieves end-to-end quadruple extraction by combining rich dialogue features and discourse features. \textbf{MRM-STS}  \citep{www_baseline} is a span-based method that includes a novel Span-pair Tagging Scheme (STS) and a simple and efficient Multi-level Representation Model (MRM). 
STS explicitly captures complete span-level semantics by tagging span pairs. 
MRM efficiently models the dialogue structure information and span-level
interactions by constructing multi-level contextual representation. \textbf{H2DT}  \citep{h2dt} is a grid-tagging-based method that constructs a token-level heterogeneous graph and enhances token interactions through a heterogeneous attention network and utilizes a novel triadic scorer, strengthening weak token relations within a quadruple, thereby enhancing the cohesion of the quadruple extraction. \textbf{Qwen3-Max} \citep{qwen3} and \textbf{GPT-5} \citep{gpt-5} are currently the state-of-the-art LLMs. 

\begin{table}[!h]
\fontsize{12}{13}\selectfont
\setlength{\tabcolsep}{0.8mm}
\setlength{\arrayrulewidth}{0.1mm}
\centering
\resizebox{0.4\textwidth}{!}{
\begin{tabular}{l c c}
\bf Hyperparameter & \phantom{} &\bf Value \\
\hline
Learning Rate (LLMs) & \phantom{} & 1e-5\\
Learning Rate (Others) & \phantom{} & 1e-3\\
Epochs &\phantom{} & 15\\
Batch Size &\phantom{} & 1\\
$h_{p}$ &\phantom{} & 20\\
$h_{s}$ &\phantom{} & 512\\
$h_{e}$ &\phantom{} & 256\\
$h_{r}$ &\phantom{} & 50\\
Dropout &\phantom{} & 0.5\\
\hline
\end{tabular}
}
\caption{Detailed hyperparameter settings.}
\label{tab:hyperparameter}
\end{table}

\subsection{Evaluation Metrics}
Our evaluation metrics follow \cite{meatwp}, using the precision (P), recall (R) and F1.
These metrics can be used to detect entities, relations and quadruple.
For an item to be considered a correct prediction, it needs to match the gold standard exactly.

\subsection{Implementation Details}
In our experiments, the backbone models of all baseline models (except ParaPhrase and LLMs) and our model were all built using \textit{XLM-RoBERTa-base} \citep{xlm_roberta}. The backbone model of ParaPhrase used \textit{mT5-base} \citep{mt5}. For LLMs, we conduct tests under zero-shot and in-context learning settings.
Our model is implemented using PyTorch and trained using NVIDIA RTX 3090 GPU.
 We used the Stanford Stanza toolkit \citep{Stanza} to extract syntactic knowledge. 
The hyperparameters are adjusted according to the fine-tuning on the development set, our hyperparameter settings are shown in Table \ref{tab:hyperparameter}.

\begin{table*}[!h]
\fontsize{12}{13}\small
\setlength{\tabcolsep}{0.8mm}
\setlength{\arrayrulewidth}{0.1mm}
\centering
\resizebox{1\textwidth}{!}{
\begin{tabular}{c l c c c c c c c c c c c c}
\hline
& & \phantom{} &\multicolumn{3}{c}{ \bf Entity (F1)}& \phantom{} & \multicolumn{3}{c}{\bf Relation (F1)} & \phantom{} & \multicolumn{3}{c}{ \bf Quadruple} \\
\cmidrule(lr){4-6}\cmidrule(lr){8-10}\cmidrule(lr){12-14}
& & \phantom{} &  T &  A & O &\phantom{} &  TA &  TO & AO &\phantom{} & P & R & F1\\
\hline
\multirow{11}{*}{\bf UZ} & EC-ACOS&\phantom{}&64.06&23.32&44.75&\phantom{}&16.41&30.92&15.16&\phantom{}&16.94&8.72&11.51\\
&Span-ASTE&\phantom{}&65.21&24.29&46.81&\phantom{}&17.92&32.17&16.83&\phantom{}&18.96&10.31&13.36\\
&ParaPhrase&\phantom{}&63.16&22.86&47.87&\phantom{}&16.91&33.50&15.89&\phantom{}&22.27&8.96&12.78\\ 
&  Meta-WP & \phantom{} &\cellcolor{bc3}75.62&\cellcolor{bc3}25.35&\cellcolor{bc3}54.51&\cellcolor{bc3}\phantom{} &\cellcolor{bc3}20.57&\cellcolor{bc3}40.40&\cellcolor{bc3}19.29&\cellcolor{bc3}\phantom{} &\cellcolor{bc3}\bf25.52&\cellcolor{bc3}11.13&\cellcolor{bc3}15.50\\
 &  MRM-STS & \phantom{} &\cellcolor{bc3}73.52&\cellcolor{bc3}26.52&\cellcolor{bc3}50.57&\cellcolor{bc3}\phantom{} &\cellcolor{bc3}21.05&\cellcolor{bc3}39.70&\cellcolor{bc3}20.26&\cellcolor{bc3}\phantom{} &\cellcolor{bc3}18.66&\cellcolor{bc3}12.82&\cellcolor{bc3}15.90\\
 &  H2DT & \phantom{} &\cellcolor{bc3}75.77&\cellcolor{bc3}25.17&\cellcolor{bc3}54.65&\cellcolor{bc3}\phantom{} &\cellcolor{bc3}19.69&\cellcolor{bc3}40.36&\cellcolor{bc3}19.96&\cellcolor{bc3}\phantom{} &\cellcolor{bc3}24.30&\cellcolor{bc3}11.76&\cellcolor{bc3}15.85\\
 \cdashline{2-14}
 &  Qwen3-Max (ZS) & \phantom{} &35.51&14.05&26.22&\phantom{}&11.01&26.01&12.31&\phantom{}&15.53&5.91&8.56\\
 &  GPT-5 (ZS)     & \phantom{} &38.32&13.89&27.75&\phantom{}&13.71&25.11&14.55&\phantom{}&19.30&6.52&9.75\\
  & Qwen3-Max (ICL) & \phantom{} &40.06&16.86&30.25&\phantom{}&14.29&30.21&15.38&\phantom{}&18.64&8.04&11.23\\
 & PT-5 (ICL)    & \phantom{} &40.24&16.56&30.53&\phantom{}&15.35&27.02&16.30&\phantom{}&20.23&9.50&12.93\\
 \cdashline{2-14}
 & \multirow{2}{*}{Our Model}    & \phantom{} &\cellcolor{bc3}\bf75.90&\cellcolor{bc3}\bf28.44&\cellcolor{bc3}\bf54.86&\cellcolor{bc3}\phantom{} &\cellcolor{bc3}\bf22.27&\cellcolor{bc3}\bf40.79&\cellcolor{bc3}\bf20.45&\cellcolor{bc3}\phantom{} &\cellcolor{bc3}22.71&\cellcolor{bc3}\bf13.82&\cellcolor{bc3}\bf17.18\\
  &  & \phantom{} &(1.54)&(0.85)&(1.20)&\cellcolor{bc3}\phantom{} &(0.76)&(1.47)&(0.41)&\cellcolor{bc3}\phantom{} &(0.55)&(1.05)&(0.97)\\
\hline
\multirow{11}{*}{\bf UY} & EC-ACOS&\phantom{}&69.32&23.81&37.55&\phantom{}&18.21&30.82&15.59&\phantom{}&14.29&8.006&10.31\\
&Span-ASTE&\phantom{}&68.31&24.76&39.29&\phantom{}&18.78&31.83&15.65&\phantom{}&15.55&9.36&11.69\\
&ParaPhrase&\phantom{}&62.65&20.95&40.26&\phantom{}&16.72&31.77&13.57&\phantom{}&18.63&7.26&10.45\\
&Meta-WP & \phantom{} &\cellcolor{bc4}79.74&\cellcolor{bc4}24.68&\cellcolor{bc4}50.25&\cellcolor{bc4}\phantom{} &\cellcolor{bc4}20.23&\cellcolor{bc4}38.45&\cellcolor{bc4}17.23&\cellcolor{bc4}\phantom{} &\cellcolor{bc4}\bf23.52&\cellcolor{bc4}9.77&\cellcolor{bc4}13.81\\
 &  MRM-STS & \phantom{} &\cellcolor{bc4}77.51&\cellcolor{bc4}26.11&\cellcolor{bc4}47.94&\cellcolor{bc4}\phantom{} &\cellcolor{bc4}21.21&\cellcolor{bc4}39.71&\cellcolor{bc4}16.68&\cellcolor{bc4}\phantom{} &\cellcolor{bc4}16.04&\cellcolor{bc4}11.80&\cellcolor{bc4}13.59\\
 &  H2DT & \phantom{} &\cellcolor{bc4}80.90&\cellcolor{bc4}24.63&\cellcolor{bc4}49.86&\cellcolor{bc4}\phantom{} &\cellcolor{bc4}20.38&\cellcolor{bc4}39.80&\cellcolor{bc4}17.31&\cellcolor{bc4}\phantom{} &\cellcolor{bc4}20.45&\cellcolor{bc4}11.05&\cellcolor{bc4}14.25\\
  \cdashline{2-14}
 &  Qwen3-Max (ZS) & \phantom{} &31.63&12.32&20.57&\phantom{}&9.87&20.59&9.54&\phantom{}&10.09&6.38&7.82\\
 &  GPT-5 (ZS)    & \phantom{} &40.79&14.41&22.83&\phantom{}&10.49&22.17&11.00&\phantom{}&12.57&6.53&8.59\\
 &  Qwen3-Max (ICL) & \phantom{} &34.21&15.80&24.63&\phantom{}&12.15&23.65&11.78&\phantom{}&12.40&8.34&9.97\\
     &  GPT-5 (ICL)     & \phantom{} &42.83&16.14&25.11&\phantom{}&11.85&24.39&12.43&\phantom{}&14.20&8.51&10.64\\
 \cdashline{2-14}
 & \multirow{2}{*}{Our Model}    & \phantom{} &\cellcolor{bc4}\bf80.46&\cellcolor{bc4}\bf27.75&\cellcolor{bc4}\bf50.66&\cellcolor{bc4}\phantom{} &\cellcolor{bc4}\bf23.63&\cellcolor{bc4}\bf40.01&\cellcolor{bc4}\bf18.28&\cellcolor{bc4}\phantom{} &\cellcolor{bc4}19.77&\cellcolor{bc4}\bf13.60&\cellcolor{bc4}\bf16.11\\
   &  & \phantom{} &(1.13)&(2.38)&(2.14)&\cellcolor{bc3}\phantom{} &(1.20)&(0.83)&(2.29)&\cellcolor{bc3}\phantom{} &(2.50)&(2.24)&(1.34)\\
\hline
\end{tabular}
}
\caption{Comparison of our method with baseline models on our LASQ dataset. T/A/O represents target, aspect and opinion, and TA/TO/AO represents target-aspect pair, target-opinion pair and aspect-opinion pair. \textbf{Bold} represents the best results for each column under different language settings.The values in parentheses represent the variance of the results under three random numbers. ZS represents the zero-shot settings, and ICL represents the in-context learning settings.}
\label{tab:main_result}
\end{table*}

\section{Results and Analysis}
\subsection{Main Comparisons}

\begin{table*}[!t]
\fontsize{12}{13}\small
\setlength{\tabcolsep}{0.8mm}
\setlength{\arrayrulewidth}{0.1mm}
\centering
\resizebox{1\textwidth}{!}{
\begin{tabular}{c l c c c c c c c c c c c c}
\hline
& & \phantom{} &\multicolumn{3}{c}{ \bf Entity (F1)}& \phantom{} & \multicolumn{3}{c}{\bf Relation (F1)} & \phantom{} & \multicolumn{3}{c}{ \bf Quadruple} \\
\cmidrule(lr){4-6}\cmidrule(lr){8-10}\cmidrule(lr){12-14}
& & \phantom{} &  T &  A & O &\phantom{} &  TA &  TO & AO &\phantom{} & P & R & F1\\
\hline
\multirow{9}{*}{\bf UZ} & Our Model   & \phantom{} &75.90&28.44&54.86&\phantom{} &22.27&40.79&20.45&\phantom{} &22.71&13.82&17.18\\
 \cdashline{2-14}
& \quad $w/o$ POS &\phantom{}&76.43 &26.56 &54.43 &\phantom{}&21.59 &40.60 &19.44 &\phantom{}&24.02 &12.54 &16.48 \\
& \quad $w/o$ Dependency&\phantom{}&75.95 &25.04 &54.42 &\phantom{}&19.77 &40.37 &18.67 &\phantom{}&22.53 &12.55 &16.12 \\
& \quad $w/o$ SKEM &\phantom{}&77.44 &26.12 &55.19 &\phantom{}&21.21 &42.41 &20.54 &\phantom{}&21.16 &12.73 &15.90\\
 \cdashline{2-14}
& \quad Random POS &\phantom{}&76.32 &24.55 &53.77 &\phantom{}&19.96 &40.67 &18.65 &\phantom{}&24.00&12.005&16.04 \\
& \quad Random Dependency &\phantom{}&76.02 &24.23 &53.94 &\phantom{}&20.01 &40.19 &18.93 &\phantom{}&24.13&11.76&15.81\\
& \quad Random SK &\phantom{}&76.62 &25.51 &54.91 &\phantom{}&21.28 &41.84 &19.23 &\phantom{}&20.19&12.19&15.20\\
 \cdashline{2-14}
& \quad $w/o$ Biaffine&\phantom{}&75.91 &26.41 &53.83 &\phantom{}&21.15 &40.33 &19.71 &\phantom{}&26.37 &12.25 &16.73 \\
& \quad $w/o$ Triaffine &\phantom{}&75.79 &23.98 &54.28 &\phantom{}&19.14 &39.91 &17.91 &\phantom{}&21.08 &13.01 &16.09\\
\hline
\multirow{9}{*}{\bf UY}  &Our Model   & \phantom{} &80.46&27.75&50.66&\phantom{} &23.63&40.01&18.28&\phantom{} &19.77&13.60&16.11\\
 \cdashline{2-14}
& \quad $w/o$ POS &\phantom{}&82.13 &24.43 &50.62 &\phantom{}&20.63 &41.27 &17.57 &\phantom{}&21.06 &12.52 &15.70\\
& \quad $w/o$ Dependency&\phantom{}&79.66 &23.92 &50.96 &\phantom{}&19.96 &39.76 &16.96 &\phantom{}&20.75 &12.02 &15.22 \\
& \quad $w/o$ SKEM &\phantom{}&81.25 &24.23 &50.09 &\phantom{}&21.07 &40.42 &17.55 &\phantom{}&20.81 &11.81 &15.07\\
 \cdashline{2-14}
& \quad Random POS &\phantom{}&80.05 &24.41 &50.19 &\phantom{}&20.55 &39.68 &17.24 &\phantom{}&19.04&11.78&14.55\\
& \quad Random Dependency &\phantom{}&79.71 &24.42 &50.50 &\phantom{}&19.98 &38.97 &17.33 &\phantom{}&19.15&11.32&14.23\\
& \quad Random SK &\phantom{}&80.24 &24.28 &51.05 &\phantom{}&20.18 &40.18 &16.47 &\phantom{}&18.69&10.66&13.58\\
 \cdashline{2-14}
& \quad $w/o$ Biaffine&\phantom{}&79.65 &25.91 &50.38 &\phantom{}&21.24 &40.04 &18.01 &\phantom{}&21.30 &12.03 &15.38  \\
& \quad $w/o$ Triaffine &\phantom{}&81.45 &22.87 &49.95 &\phantom{}&19.18 &40.27 &15.16 &\phantom{}&20.79 &11.58 &14.87\\

\hline
\end{tabular}
}
\caption{Ablation experiment.}
\label{tab:Ablation}
\end{table*}

Our main experimental results are shown in Table \ref{tab:main_result}. Regarding the quadruple extraction results, our model achieved optimal performance in both Uzbek and Uyghur languages, with F1 scores improved by 1.28\% (17.18\% - 15.90\%) and 1.86\% (16.11\% - 14.25\%) respectively compared to the best baseline model. This is because our model explicitly incorporates part-of-speech and syntactic dependency knowledge, enabling more accurate extraction of entity words and their relations. The extraction results for both entities and relations show improvements in all three types of entity words and relations, which is a direct result of the improved quadruple extraction. 
Furthermore, we compared our model with two powerful LLMs. The results show that, in the zero-shot setting, Qwen3-Max \citep{qwen3} and GPT-5 \citep{gpt-5} performed poorly, even worse than earlier methods (EC-ACOS). Compared to our method, they differed by at least 7.43\% (17.18\% - 9.75\%) and 7.52\% (16.11\% - 8.59\%) in F1 scores for quadruple extraction in Uzbek and Uyghur languages. Even with ICL, there is still a gap compared to our model. This performance deficiency is not only due to the inherent limitations of large models in handling fine-grained tasks, coupled with a lack of fine-tuning leading to severe illusion problems, but also because existing LLMs lack a deep understanding of the vocabulary, grammar, and semantics of these two languages, thus failing to effectively extract sentiment quadruples.


\subsection{Ablation Experiment}

We also conducted ablation experiments, and the results are shown in Table \ref{tab:Ablation}. We first analyzed the effectiveness of our proposed SKEM, including removing the entire SKEM and removing POS and dependency knowledge separately from the SKEM. Regardless of the variant, the model performance declined. Removing the entire SKEM naturally resulted in the largest performance drop, with decreases of 1.29\% (17.18\% - 15.90\%) and 1.05\% (16.11\% - 15.07\%) in Uzbek and Uyghur respectively. Furthermore, we found that removing dependency knowledge caused a greater performance drop than removing POS knowledge. This may be because dependency knowledge characterizes the relations between words (such as subject-verb, verb-object, and modifying relations), which is a global, structured form of knowledge, more useful than POS, which describes part-of-speech categories at the local, lexical level. 
To more accurately verify the effectiveness of the knowledge we introduced, we no longer removed the modules, but instead directly used randomly generated knowledge. Here, Random POS and Random Dependency represent completely randomly generated POS labels and dependencies, while Random SK contains Random POS and Random Dependency. Experimental results show that randomly generated knowledge leads to a greater performance drop in the model compared to removing the corresponding modules. However, this also demonstrates that our model can extract useful information from this syntactic knowledge, thereby helping the model identify quadruples.
Finally, we removed biaffine and triaffine, and the model performance also decreased significantly, indicating their effectiveness.

\begin{figure*}[!h]
\centering
\includegraphics[width=0.8\textwidth]{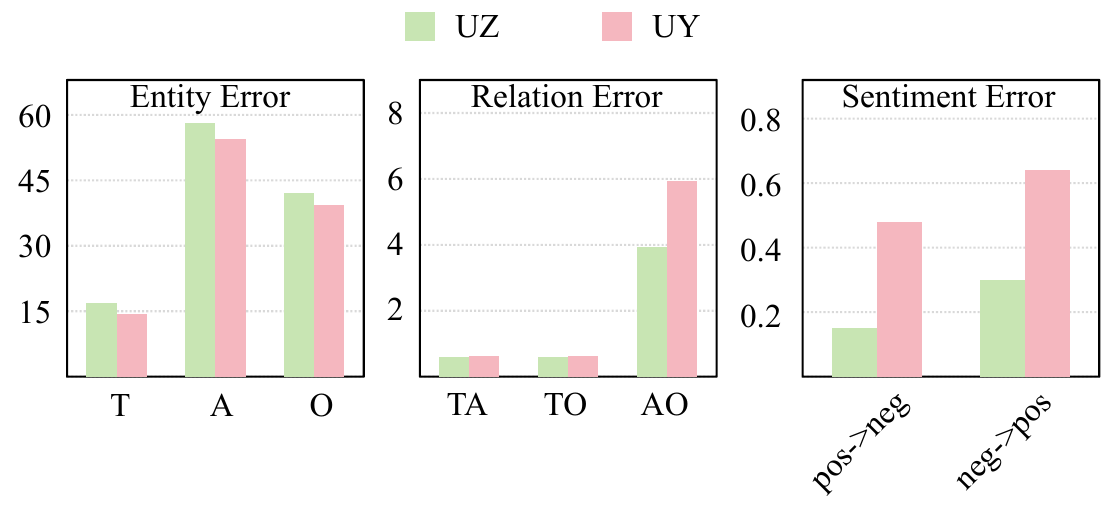}
\caption{Our method performs error analysis on quadruple extraction.
} 
\label{error_quad}
\end{figure*}
\subsection{Error Analysis}
We performed error analysis\footnote{Error rate for each error type = number of quadruples with that error type / total number of quadruples} on our method, categorizing the error types of the predicted quadruples into three main categories: entity errors, relation errors and sentiment errors. Entity errors were further subdivided into three subcategories: target errors (T), aspect errors (A) and opinion errors (O). Relation errors were also subdivided into three subcategories: target-aspect relation errors (TA), target-opinion relation errors (TO) and aspect-opinion relation errors (AO). Sentiment errors were subdivided into two subcategories: false negative errors (pos->neg) and false positive errors (neg->pos). The specific error analysis results are shown in Figure \ref{error_quad}.

The results show that on both language datasets, the main errors almost exclusively stemmed from entity errors, with aspect identification errors being the most frequent. This is likely because aspect diversity is far greater than that of target and opinion entities. Furthermore, we found that AO errors were significantly higher than TO and TA errors in relational errors. This is natural, as the high error rates of aspect and opinion entities directly lead to a high error rate in AO relations. Finally, we observed slightly more false positive errors than false negative errors, which may be due to data imbalance.

\subsection{Case Studies}

\begin{table}[t]
\small
\centering
\caption{Case study on Uzbek reviews about Google applications: effect of syntactic knowledge.}
\begin{tabular}{>{\raggedright\arraybackslash}p{0.95\linewidth}}
\toprule
\textbf{Case 1 (Gmail)}: \textit{Gmail interfeysi qulay, lekin sinxronlash ba’zan sekin.}\\
Gold: (Gmail, interfeys, qulay, pos); (Gmail, sinxronlash, sekin, neg)\\
H2DT: (Gmail, interfeys, qulay, pos); \uline{(Gmail, interfeys, sekin, neg)}\\
Ours: (Gmail, interfeys, qulay, pos); (Gmail, sinxronlash, sekin, neg)\\
\midrule
\textbf{Case 2 (Google Maps)}: \textit{Google Maps yo‘nalishlarni aniq ko‘rsatadi, ammo ovozli navigatsiya tez-tez uzilib qoladi.}\\
Gold: (Google Maps, yo‘nalishlar, aniq, pos); (Google Maps, ovozli navigatsiya, uzilib qoladi, neg)\\
H2DT: (Google Maps, yo‘nalishlar, aniq, pos); \uline{(Google Maps, yo‘nalishlar, uzilib qoladi, neg)}\\
Ours: (Google Maps, yo‘nalishlar, aniq, pos); (Google Maps, ovozli navigatsiya, uzilib qoladi, neg)\\
\bottomrule
\label{tab:lizi}
\end{tabular}
\end{table}
We further present a case study on Uzbek reviews about Google applications to demonstrate the benefit of injecting syntactic knowledge. As shown in Table~\ref{tab:lizi}, in both examples, the syntax-agnostic baseline H2DT exhibits aspect–opinion misalignment under adversative constructions (e.g., \textit{lekin}/\textit{ammo} “but”), where negative opinions in the second clause are incorrectly attached to an earlier aspect due to surface proximity (e.g., linking “slow” to “interface” instead of “synchronization” in Gmail). By contrast, our syntax-aware model leverages dependency-based cues to align opinion expressions with their syntactically governed targets (predicate–argument or modifier–head relations), thereby producing correct quadruple extraction. These qualitative results highlight that syntactic priors are particularly effective for mitigating cross-clause mismatches and improving robustness under long-range dependencies.

\subsection{Efficiency Analysis}

\begin{table}[!t]
\fontsize{11}{13}\selectfont
\setlength{\tabcolsep}{0.8mm}
\centering
\resizebox{0.45\textwidth}{!}{
\begin{tabular}{c l c c c c}
\hline
& & \phantom{}&\makecell{\bf Training  }& \phantom{} &\makecell{\bf Inference} \\
& & \phantom{}&\makecell{  \bf(doc/s)}& \phantom{} &\makecell{ \bf (doc/s)} \\
\hline
\multirow{7}{*}{\bf UZ} & EC-ACOS &\phantom{}&5.60&\phantom{}&14.35\\
& Span-ASTE &\phantom{}&5.15&\phantom{}&14.79\\
&ParaPhrase &\phantom{}&10.36&\phantom{}&9.74\\
& Meta-WP &\phantom{}&21.31&\phantom{}&76.50\\
&  MRM-STS &\phantom{}&5.06&\phantom{}&13.30\\
&  H2DT&\phantom{} &12.38&\phantom{}&23.54\\
\cdashline{2-6}
& Our Model &\phantom{} &13.93 & \phantom{} &27.82\\
\hline
\multirow{7}{*}{\bf UY} & EC-ACOS &\phantom{}&5.53&\phantom{}&10.98\\
& Span-ASTE &\phantom{}&5.20&\phantom{}&11.43\\
&ParaPhrase &\phantom{}&10.71&\phantom{}&8.25\\
& Meta-WP &\phantom{}&20.70&\phantom{}&60.40\\
&  MRM-STS &\phantom{}&4.50&\phantom{}&11.19\\
&  H2DT&\phantom{} &11.70&\phantom{}&18.88\\
 \cdashline{2-6}
& Our Model &\phantom{} &13.23 & \phantom{} & 21.57\\
\hline
\end{tabular}
}
\caption{Efficiency analysis of our model and baseline models. ``doc/s'' represents the number of sentences processed per second.
}
\label{tab:Efficiency}
\end{table}

\begingroup
We conducted an efficiency analysis of our method, and the results are shown in Table \ref{tab:Efficiency}. Our method achieved suboptimal efficiency on both languages, only worse than the Meta-WP model. This may be due to the increased computational overhead brought by the introduction of biaffine and triaffine. Compared to H2DT, our method is more efficient because our syntactic structure is a homomorphic graph, which has lower computational overhead than heteromorphic graphs. Furthermore, we noted that span-based methods (such as EC-ACOS, Span-ASTE and MRM-STS) and generative methods (such as ParaPhrase) are both inefficient. This is mainly because span-based methods require enumerating spans, while generative models are inefficient due to their autoregressive generation mechanism that produces characters sequentially.

{
\begin{figure*}[!t]
\centering
\includegraphics[width=0.6\textwidth]{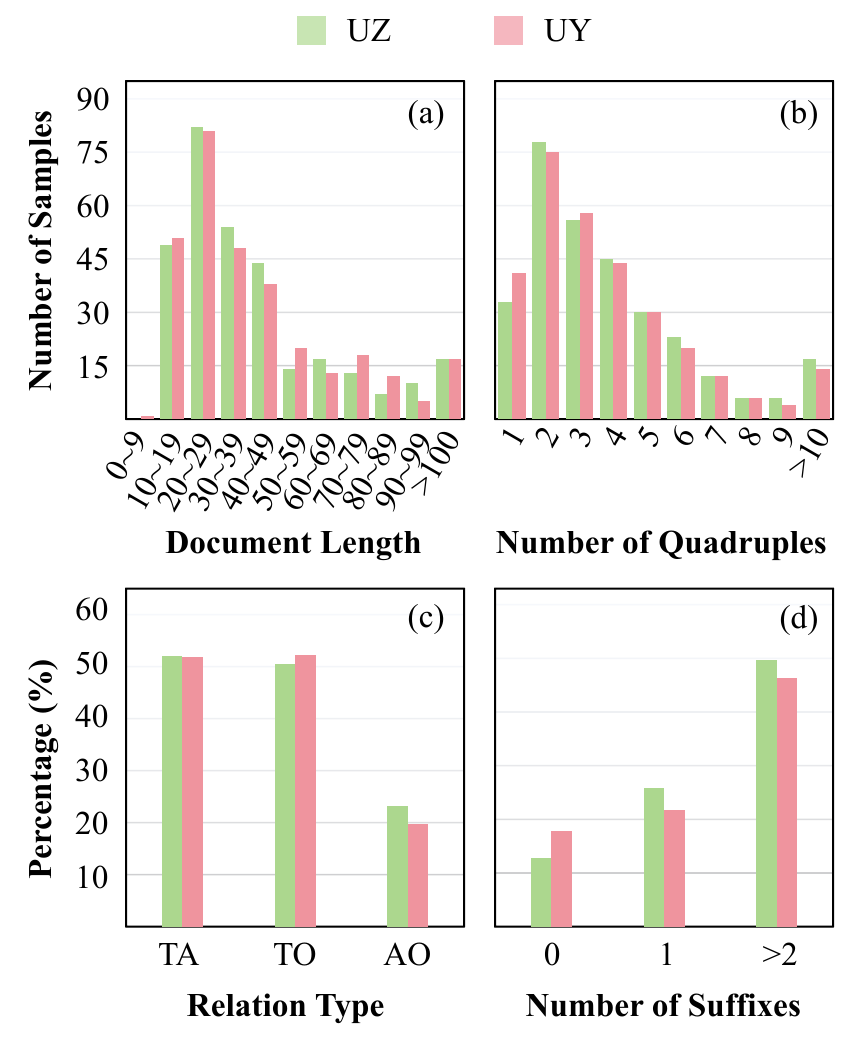}
\caption{We present the results of a more in-depth analysis of our LASQ dataset. Figure (a) shows the statistics for document length; Figure (b) shows the statistics for quadruples; Figure (c) shows the proportion of cross-sentence relations; and Figure (d) shows the proportion of entities with different morphological complexities.
} 
\label{indepth}
\end{figure*}

\section{In-Depth Analysis of the LASQ Dataset}
We conducted a more in-depth analysis of our LASQ dataset, and the specific results are shown in Figure \ref{indepth}.

\noindent\textit{\textbf{Document Length:}} We divided the document length into 11 groups and then counted the number of samples contained in each group, as shown in Figure \ref{indepth}(a). We found that the document length of most samples in both languages is concentrated between 10 and 49, and even the longest document is less than 140. For existing pre-trained language models, their self-attention mechanism can capture the associations between all words without hindrance, and there is no problem of "long-distance dependency loss".

\noindent\textit{\textbf{Number of Quadruples:}} We divided the number of quadruples into 10 groups and then counted the number of samples in each group. The results are shown in Figure \ref{indepth}(b). We found that compared with other low-resource ABSA datasets \cite{uzabsa,Vietnamese} (which typically contain at most three quadruples), our dataset has nearly half of the samples containing at least four quadruples in both languages, and some even contain more than ten quadruples. This is mainly because our dataset is document-level, and longer texts usually have more quadruples. However, this also poses a challenge to the model's recognition. Specifically, 1) more quadruples indicate more entities such as targets, aspects, and opinions, which makes it easier for the model to miss some entities; 2) more entities lead to more complex relations between them, which manifests as challenges for the model in recognizing TA, TO, and AO relations.

\noindent\textit{\textbf{Cross-Sentence Relations:}} Figure \ref{indepth}(c) shows the proportion of cross-sentence relations for the three types of relations. We found that relations related to the target entity have a high proportion of cross-sentence relations. This is not a characteristic of low-resource languages, but rather a domain characteristic of our data. Our data comes from user reviews of products, and such data usually only mentions the target entity at the beginning of the review and then omits it, resulting in half of the TA and TO relations being cross-sentence relations. However, cross-sentence relations did not have a serious impact on our model, as the results in Table \ref{tab:main_result} show that the recognition F1 score of the TO relation with a high proportion of cross-sentence relations is much higher than that of the AO relation with a low proportion of cross-sentence relations.

\noindent\textit{\textbf{Morphological Complexity:}} We analyzed the morphological complexity of entity words, dividing them into three groups according to the number of suffixes they contain, as shown in Figure \ref{indepth}(d). Compared to English, which typically contains at most one suffix, we believe that words containing at least two suffixes have higher morphological complexity. We found that about half of the entity words in both languages have high complexity, which causes the model to split these words into multiple tokens, making it difficult to accurately identify entity boundaries during the final recognition.

\section{Future Work}
In this study, we propose an initial method for our LASQ dataset. While its performance surpasses existing baseline models, further improvements are possible from several angles. To facilitate future research in this direction, we attempt to elucidate several potential directions for future work.

\noindent\textit{\textbf{Exploiting the Semantic Reasoning Capabilities of Large Language Models.}} Compared to static embedding-based methods, large language models (MLMs) offer superior contextual understanding and logical reasoning capabilities. Specifically: 1) As a backbone network, suitable prompt templates can be designed to fine-tune LLMs, either fully or based on LoRA. Given that the ABSA task involves extracting structured sentiment quadruples, we speculate that code-style prompts (e.g., formatting output as JSON or Python objects) will be more effective. 2) As an external knowledge base, large language models can serve as a rich source of implicit world knowledge and common-sense reasoning, generating high-quality synthetic data for data augmentation or providing explicit reasoning chains to explain why specific sentiment polarities are assigned to certain aspects.

\noindent\textit{\textbf{Enhancing Aspect Term Detection under High Lexical Diversity.}} The LASQ dataset exhibits a high degree of diversity in aspects, posing a challenge to our initial model in recognizing them. We propose two possible improvements: 1) aspect-oriented data augmentation, utilizing generative models to generate synthetic training samples containing diverse aspect paraphrases to enrich the training data distribution; and 2) introducing an external knowledge base to explicitly link different surface forms to a unified semantic concept. By combining these strategies, a more robust representation space can be constructed to capture the full scope of aspect terminology diversity.

\noindent\textit{\textbf{Optimizing Computational Efficiency of High-Order Interaction Scorers.}} Although biaffine and triaffine scorers can achieve rich interactions between word pairs and triples, their quadratic and cubic computational complexity reduces the efficiency of our initial model. Therefore, we plan to explore: (1) low-rank decomposition, which approximates the complete interaction tensor by decomposing it into smaller low-rank matrices, thereby significantly reducing the number of parameters and floating-point operations; and (2) parameter sharing strategies, which share scoring weights across different interaction layers or attention head dimensions to further compress the model size without sacrificing expressive power. By integrating these compression mechanisms, a favorable balance between computational efficiency and prediction accuracy can be achieved.

\noindent\textit{\textbf{Mitigating Error Propagation from External Syntactic Parsers.}} Our current process relies on readily available part-of-speech taggers and dependency parsers to guide quadruple extraction—a design choice that can lead to error propagation when syntactic tagging is noisy or mismatched with domain-specific syntax. We can consider: 1) Introducing uncertainty estimation and confidence filtering, i.e., adding a confidence score to the parser's output. Edges or nodes with confidence scores below a threshold are not directly used for subsequent analysis or subjected to secondary processing. 2) Utilizing large language models for self-correction, scoring or reconstructing the generated top-n candidate trees using LLMs to obtain the tree that best conforms to semantic logic.

\noindent\textit{\textbf{Reducing Extraction Errors in Multi-Quadruple Scenarios.}} Nearly half of the samples in our dataset contain at least four quadruplets, and some even contain more than ten quadruplets. This makes the entities denser and the relations more complex, making it easier for the model to miss or make errors in identifying entities and relations. We believe that a multi-round iterative extraction framework can be designed. Specifically, in each round of model recognition, we mask the entities or relations identified in previous rounds, thereby ensuring that as many new entities and relations as possible are identified and avoiding omissions.

\noindent\textit{\textbf{Tackling the Impact of High Morphological Complexity.}} Uzbek and Uyghur, as agglutinative languages, have higher morphological complexity compared to English. In our dataset, half of the entity words have at least two suffixes. This high morphological complexity causes the model to segment entity words into multiple words, making it difficult to determine entity word boundaries. Therefore, we believe future models could: 1) perform more fine-grained modeling at the character or sub-word level, for example, using character-level encoders to extract morphological features within words and use them as supplementary information for pre-trained model input; or 2) further enhance our SKEM, specifically by incorporating morphological features such as case markers and person agreement as additional syntactic information into the model.
}
\section{CONCLUSION}\label{sec:7}

This paper constructs a high-quality low-resource language ABSA dataset, named LASQ, which includes two languages: Uzbek and Uyghur. We then design a grid-tagging-based model for low-resource languages. Through a designed syntax knowledge embedding module (SKEM), we integrate POS and syntactic knowledge into the model, thereby alleviating the sparsity problem caused by morphological variations in these two agglutinative languages. Experimental results demonstrate that our method outperforms several existing baseline models, proving its effectiveness. Furthermore, we conducted more extensive experimental analyses on our dataset and model. Finally, we discuss potential future research directions, including issues related to the diversity of aspect terms, morphological complexity, and efficiency. We hope our work will advance research in the field of low-resource language ABSA and attract more scholars to this area.

\noindent\textbf{Acknowledgements.} This work was supported by the Xinjiang Uygur Autonomous Region Innovation Environment (Talent and Base) Construction Special Program---Natural Science Program (Special Training for Ethnic Minority Scientific and Technological Talents) (Grant No.~2025D03032), the Youth Top-notch Talent Support Program of Xinjiang Normal University (Grant No.~XJNUQB2022-22), the Tender Project of the Engineering Research Center of Smart Education of Xinjiang Normal University (Grant No.~XJNU-ZHJY202403), the National Natural Science Foundation of China (Grant No.~61662081), and the National Social Science Foundation of China (Grant No.~14AZD11).






\bibliography{sn-bibliography}%

@inproceedings{ABSA1,
  title={Multi-entity aspect-based sentiment analysis with context, entity and aspect memory},
  author={Yang, Jun and Yang, Runqi and Wang, Chongjun and Xie, Junyuan},
  booktitle={Proceedings of the AAAI Conference on Artificial Intelligence},
  volume={32},
  number={1},
  year={2018}
}

@inproceedings{Stanza,
    title={Stanza: A {Python} Natural Language Processing Toolkit for Many Human Languages},
    author={Qi, Peng and Zhang, Yuhao and Zhang, Yuhui and Bolton, Jason and Manning, Christopher D.},
    booktitle = "Proceedings of the 58th Annual Meeting of the Association for Computational Linguistics: System Demonstrations",
    year={2020}
}

@article{syntactic_3,
  author={Wu, Chunling and Kang, Houwei}, 
  title={Harnessing Syntax GCN and Multi-View Interaction for Conversational Aspect-Based Quadruple Sentiment Analysis}, 
  year={2025},
  volume={13},
  pages={73332-73341},
}

@article{syntactic_2,
  title={Modeling implicit variable and latent structure for aspect-based sentiment quadruple extraction},
  author={Nie, Yu and Fu, Jianming and Zhang, Yilai and Li, Chao},
  journal={Neurocomputing},
  volume={586},
  pages={127642},
  year={2024},
}

@article{syntactic_1,
  title={Block-level dependency syntax based model for end-to-end aspect-based sentiment analysis},
  author={Xiang, Yan and Zhang, Jiqun and Guo, Junjun},
  journal={Neural Networks},
  volume={166},
  pages={225--235},
  year={2023}
}

@inproceedings{mams,
  title={A challenge dataset and effective models for aspect-based sentiment analysis},
  author={Jiang, Qingnan and Chen, Lei and Xu, Ruifeng and Ao, Xiang and Yang, Min},
  booktitle={Proceedings of the 2019 conference on empirical methods in natural language processing and the 9th international joint conference on natural language processing (EMNLP-IJCNLP)},
  pages={6280--6285},
  year={2019}
}

@inproceedings{aste,
  title={Knowing what, how and why: A near complete solution for aspect-based sentiment analysis},
  author={Peng, Haiyun and Xu, Lu and Bing, Lidong and Huang, Fei and Lu, Wei and Si, Luo},
  booktitle={Proceedings of the AAAI conference on artificial intelligence},
  volume={34},
  number={05},
  pages={8600--8607},
  year={2020}
}

@inproceedings{uzabsa,
  title={UzABSA: Aspect-Based Sentiment Analysis for the Uzbek Language},
  author={Matlatipov, Sanatbek Gayratovich and Rajabov, Jaloliddin and Kuriyozov, Elmurod and Aripov, Mersaid},
  booktitle={Proceedings of the 3rd Annual Meeting of the Special Interest Group on Under-resourced Languages@ LREC-COLING 2024},
  pages={394--403},
  year={2024}
}

@inproceedings{xlm_roberta,
  title={Unsupervised Cross-lingual Representation Learning at Scale},
  author={Conneau, Alexis and Khandelwal, Kartikay and Goyal, Naman and Chaudhary, Vishrav and Wenzek, Guillaume and Guzm{\'a}n, Francisco and Grave, {\'E}douard and Ott, Myle and Zettlemoyer, Luke and Stoyanov, Veselin},
  booktitle={Proceedings of the 58th Annual Meeting of the Association for Computational Linguistics},
  pages={8440--8451},
  year={2020}
}

@article{roberta,
  title={Roberta: A robustly optimized bert pretraining approach},
  author={Liu, Yinhan and Ott, Myle and Goyal, Naman and Du, Jingfei and Joshi, Mandar and Chen, Danqi and Levy, Omer and Lewis, Mike and Zettlemoyer, Luke and Stoyanov, Veselin},
  journal={arXiv preprint arXiv:1907.11692},
  year={2019}
}

@inproceedings{www_baseline,
  title={Span-Pair Interaction and Tagging for Dialogue-Level Aspect-Based Sentiment Quadruple Analysis},
  author={Zhou, Changzhi and Wu, Zhijing and Song, Dandan and Hu, Linmei and Tian, Yuhang and Xu, Jing},
  booktitle={Proceedings of the ACM on Web Conference 2024},
  pages={3995--4005},
  year={2024}
}

@inproceedings{h2dt,
  title={Harnessing Holistic Discourse Features and Triadic Interaction for Sentiment Quadruple Extraction in Dialogues},
  author={Li, Bobo and Fei, Hao and Liao, Lizi and Zhao, Yu and Su, Fangfang and Li, Fei and Ji, Donghong},
  booktitle={Proceedings of the AAAI Conference on Artificial Intelligence},
  volume={38},
  number={16},
  pages={18462--18470},
  year={2024}
}

@inproceedings{meatwp,
  title={DiaASQ: A Benchmark of Conversational Aspect-based Sentiment Quadruple Analysis},
  author={Li, Bobo and Fei, Hao and Li, Fei and Wu, Yuhan and Zhang, Jinsong and Wu, Shengqiong and Li, Jingye and Liu, Yijiang and Liao, Lizi and Chua, Tat-Seng and others},
  booktitle={Findings of the Association for Computational Linguistics: ACL 2023},
  pages={13449--13467},
  year={2023}
}

@inproceedings{triaffine,
  title={Efficient Second-Order TreeCRF for Neural Dependency Parsing},
  author={Zhang, Yu and Li, Zhenghua and Zhang, Min},
  booktitle={Proceedings of the 58th Annual Meeting of the Association for Computational Linguistics},
  pages={3295--3305},
  year={2020}
}

@inproceedings{Biaffine,
  title={Deep Biaffine Attention for Neural Dependency Parsing},
  author={Dozat, Timothy and Manning, Christopher D},
  booktitle={International Conference on Learning Representations},
  year={2022}
}

@inproceedings{gcn,
  title={Semi-Supervised Classification with Graph Convolutional Networks},
  author={Kipf, Thomas N and Welling, Max},
  booktitle={International Conference on Learning Representations},
  pages={1-14},
  year={2016}
}

@inproceedings{ABSA2,
  title={A unified model for opinion target extraction and target sentiment prediction},
  author={Li, Xin and Bing, Lidong and Li, Piji and Lam, Wai},
  booktitle={Proceedings of the AAAI conference on artificial intelligence},
  volume={33},
  number={01},
  pages={6714--6721},
  year={2019}
}

@book{RW1,
  title={Sentiment analysis: Mining opinions, sentiments, and emotions},
  author={Liu, Bing},
  year={2020},
  publisher={Cambridge university press}
}

@article{RW2,
  title={Sentiment analysis using deep learning architectures: a review},
  author={Yadav, Ashima and Vishwakarma, Dinesh Kumar},
  journal={Artificial Intelligence Review},
  volume={53},
  number={6},
  pages={4335--4385},
  year={2020},
  publisher={Springer}
}

@inproceedings{RW3,
author = {Pearce, Kate and Alghowinem, Sharifa and Breazeal, Cynthia},
title = {Build-a-bot: teaching conversational ai using a transformer-based intent recognition and question answering architecture},
year = {2023},
isbn = {978-1-57735-880-0},
publisher = {AAAI Press},
url = {https://doi.org/10.1609/aaai.v37i13.26903},
doi = {10.1609/aaai.v37i13.26903},
booktitle = {Proceedings of the Thirty-Seventh AAAI Conference on Artificial Intelligence and Thirty-Fifth Conference on Innovative Applications of Artificial Intelligence and Thirteenth Symposium on Educational Advances in Artificial Intelligence},
articleno = {1834},
numpages = {8},
series = {AAAI'23/IAAI'23/EAAI'23}
}

@article{RW4,
  title={Understanding the impact of control levels over emotion-aware chatbots},
  author={Benke, Ivo and Gnewuch, Ulrich and Maedche, Alexander},
  journal={Computers in Human Behavior},
  volume={129},
  pages={107122},
  year={2022},
  publisher={Elsevier}
}

@article{rw5,
  title={RETRACTED ARTICLE: Customer centric hybrid recommendation system for E-Commerce applications by integrating hybrid sentiment analysis},
  author={Karn, Arodh Lal and Karna, Rakshha Kumari and Kondamudi, Bhavana Raj and Bagale, Girish and Pustokhin, Denis A and Pustokhina, Irina V and Sengan, Sudhakar},
  journal={Electronic commerce research},
  volume={23},
  number={1},
  pages={279--314},
  year={2023},
  publisher={Springer}
}

@article{rw6,
  title={Weighted aspect-based opinion mining using deep learning for recommender system},
  author={Da'u, Aminu and Salim, Naomie and Rabiu, Idris and Osman, Akram},
  journal={Expert Systems with Applications},
  volume={140},
  pages={112871},
  year={2020},
  publisher={Elsevier}
}

@inproceedings{RW7,
  title={Self question-answering: Aspect-based sentiment analysis by role flipped machine reading comprehension},
  author={Yu, Guoxin and Li, Jiwei and Luo, Ling and Meng, Yuxian and Ao, Xiang and He, Qing},
  booktitle={Findings of the Association for Computational Linguistics: EMNLP 2021},
  pages={1331--1342},
  year={2021}
}

@article{RW8,
  title={HACAN: a hierarchical answer-aware and context-aware network for question generation},
  author={Sun, Ruijun and Tao, Hanqin and Chen, Yanmin and Liu, Qi},
  journal={Frontiers of Computer Science},
  volume={18},
  number={5},
  pages={185321},
  year={2024},
  publisher={Springer}
}

@inproceedings{RW9,
    title = "Effective {LSTM}s for Target-Dependent Sentiment Classification",
    author = "Tang, Duyu  and
      Qin, Bing  and
      Feng, Xiaocheng  and
      Liu, Ting",
    editor = "Matsumoto, Yuji  and
      Prasad, Rashmi",
    booktitle = "Proceedings of {COLING} 2016, the 26th International Conference on Computational Linguistics: Technical Papers",
    month = dec,
    year = "2016",
    address = "Osaka, Japan",
    publisher = "The COLING 2016 Organizing Committee",
    url = "https://aclanthology.org/C16-1311",
    pages = "3298--3307",
}

@article{RW10,
  title={Aspect based sentiment analysis using deep learning approaches: A survey},
  author={Chauhan, Ganpat Singh and Nahta, Ravi and Meena, Yogesh Kumar and Gopalani, Dinesh},
  journal={Computer Science Review},
  volume={49},
  pages={100576},
  year={2023},
  publisher={Elsevier}
}

@article{RW11,
  title={Deep learning for aspect-based sentiment analysis: a review},
  author={Zhu, Linan and Xu, Minhao and Bao, Yinwei and Xu, Yifei and Kong, Xiangjie},
  journal={PeerJ Computer Science},
  volume={8},
  pages={e1044},
  year={2022},
  publisher={PeerJ Inc.}
}

@INPROCEEDINGS{Indonesian,
  author={Azhar, Annisa Nurul and Khodra, Masayu Leylia},
  booktitle={2020 7th International Conference on Advance Informatics: Concepts, Theory and Applications (ICAICTA)}, 
  title={Fine-tuning Pretrained Multilingual BERT Model for Indonesian Aspect-based Sentiment Analysis}, 
  year={2020},
  volume={},
  number={},
  pages={1-6},
  doi={10.1109/ICAICTA49861.2020.9428882}}

@article{Urdu,
author = {Aziz, Kamran and Yusufu, Aizihaierjiang and Zhou, Jun and Ji, Donghong and Iqbal, Muhammad Shahid and Wang, Shijie and Hadi, Hassan Jalil and Yuan, Zhengming},
title = {UrduAspectNet: Fusing Transformers and Dual GCN for Urdu Aspect-Based Sentiment Detection},
year = {2024},
publisher = {Association for Computing Machinery},
address = {New York, NY, USA},
issn = {2375-4699},
url = {https://doi.org/10.1145/3663367},
doi = {10.1145/3663367},
note = {Just Accepted},
journal = {ACM Trans. Asian Low-Resour. Lang. Inf. Process.},
month = {may}
}

@article{Vietnamese,
author = {Van Thin, Dang and Hao, Duong Ngoc and Nguyen, Ngan Luu-Thuy},
title = {A Systematic Literature Review on Vietnamese Aspect-based Sentiment Analysis},
year = {2023},
issue_date = {August 2023},
publisher = {Association for Computing Machinery},
address = {New York, NY, USA},
volume = {22},
number = {8},
issn = {2375-4699},
url = {https://doi.org/10.1145/3610226},
doi = {10.1145/3610226},
journal = {ACM Trans. Asian Low-Resour. Lang. Inf. Process.},
month = {aug},
articleno = {218},
numpages = {28}
}

@inproceedings{Telugu,
    title = "Dataset Creation and Evaluation of Aspect Based Sentiment Analysis in {T}elugu, a Low Resource Language",
    author = "Regatte, Yashwanth Reddy  and
      Gangula, Rama Rohit Reddy  and
      Mamidi, Radhika",
    editor = "Calzolari, Nicoletta  and
      B{\'e}chet, Fr{\'e}d{\'e}ric  and
      Blache, Philippe  and
      Choukri, Khalid  and
      Cieri, Christopher  and
      Declerck, Thierry  and
      Goggi, Sara  and
      Isahara, Hitoshi  and
      Maegaard, Bente  and
      Mariani, Joseph  and
      Mazo, H{\'e}l{\`e}ne  and
      Moreno, Asuncion  and
      Odijk, Jan  and
      Piperidis, Stelios",
    booktitle = "Proceedings of the Twelfth Language Resources and Evaluation Conference",
    month = may,
    year = "2020",
    address = "Marseille, France",
    publisher = "European Language Resources Association",
    url = "https://aclanthology.org/2020.lrec-1.617",
    pages = "5017--5024",
    language = "English",
    ISBN = "979-10-95546-34-4",
}

@article{sa2,
  title={High dimensional data classification and feature selection using support vector machines},
  author={Ghaddar, Bissan and Naoum-Sawaya, Joe},
  journal={European Journal of Operational Research},
  volume={265},
  number={3},
  pages={993--1004},
  year={2018},
  publisher={Elsevier}
}

@inproceedings{sa3,
  title={Learning extraction patterns for subjective expressions},
  author={Riloff, Ellen and Wiebe, Janyce},
  booktitle={Proceedings of the 2003 conference on Empirical methods in natural language processing},
  pages={105--112},
  year={2003}
}

@inproceedings{EN2,
  title={Gti at semeval-2016 task 5: Svm and crf for aspect detection and unsupervised aspect-based sentiment analysis},
  author={Alvarez-L{\'o}pez, Tamara and Juncal-Mart{\'\i}nez, Jonathan and Fern{\'a}ndez-Gavilanes, Milagros and Costa-Montenegro, Enrique and Gonz{\'a}lez-Castano, Francisco Javier},
  booktitle={Proceedings of the 10th international workshop on semantic evaluation (SemEval-2016)},
  pages={306--311},
  year={2016}
}

@inproceedings{cnn,
  title={Aspect-based sentiment analysis approach with CNN},
  author={Mulyo, Budi M and Widyantoro, Dwi H},
  booktitle={2018 5th International Conference on Electrical Engineering, Computer Science and Informatics (EECSI)},
  pages={142--147},
  year={2018},
  organization={IEEE}
}

@article{lstm,
  title={Earlier attention? aspect-aware LSTM for aspect-based sentiment analysis},
  author={Xing, Bowen and Liao, Lejian and Song, Dandan and Wang, Jingang and Zhang, Fuzheng and Wang, Zhongyuan and Huang, Heyan},
  journal={arXiv preprint arXiv:1905.07719},
  year={2019}
}

@inproceedings{ZH1,
  title={Aspect-level sentiment analysis via convolution over dependency tree},
  author={Sun, Kai and Zhang, Richong and Mensah, Samuel and Mao, Yongyi and Liu, Xudong},
  booktitle={Proceedings of the 2019 conference on empirical methods in natural language processing and the 9th international joint conference on natural language processing (EMNLP-IJCNLP)},
  pages={5679--5688},
  year={2019}
}

@inproceedings{ZH2,
  title={Aspect-based sentiment analysis using tree kernel based relation extraction},
  author={Nguyen, Thien Hai and Shirai, Kiyoaki},
  booktitle={Computational Linguistics and Intelligent Text Processing: 16th International Conference, CICLing 2015, Cairo, Egypt, April 14-20, 2015, Proceedings, Part II 16},
  pages={114--125},
  year={2015},
  organization={Springer}
}

@article{ZH3,
  title={Attentional encoder network for targeted sentiment classification. arXiv 2019},
  author={Song, Y and Wang, J and Jiang, T and Liu, Z and Rao, Y},
  journal={arXiv preprint arXiv:1902.09314},
  year={2019}
}

@article{ZH4,
  title={Interactive attention networks for aspect-level sentiment classification},
  author={Ma, Dehong and Li, Sujian and Zhang, Xiaodong and Wang, Houfeng},
  journal={arXiv preprint arXiv:1709.00893},
  year={2017}
}

@article{Bert,
  title={Bert: Pre-training of deep bidirectional transformers for language understanding},
  author={Devlin, Jacob},
  journal={arXiv preprint arXiv:1810.04805},
  year={2018}
}

@inproceedings{ernie,
  title={Ernie 2.0: A continual pre-training framework for language understanding},
  author={Sun, Yu and Wang, Shuohuan and Li, Yukun and Feng, Shikun and Tian, Hao and Wu, Hua and Wang, Haifeng},
  booktitle={Proceedings of the AAAI conference on artificial intelligence},
  volume={34},
  number={05},
  pages={8968--8975},
  year={2020}
}

@article{liaoRoBerta,
  title={An improved aspect-category sentiment analysis model for text sentiment analysis based on RoBERTa},
  author={Liao, Wenxiong and Zeng, Bi and Yin, Xiuwen and Wei, Pengfei},
  journal={Applied Intelligence},
  volume={51},
  pages={3522--3533},
  year={2021},
  publisher={Springer}
}

@inproceedings{EN4,
  title={An Efficient Fine-tuning of Generative Language Model for Aspect-Based Sentiment Analysis},
  author={Lee, Chaelyn and Lee, Hanyong and Kim, Kyumin and Kim, Sojeong and Lee, Jaesung},
  booktitle={2024 IEEE International Conference on Consumer Electronics (ICCE)},
  pages={1--4},
  year={2024},
  organization={IEEE}
}

@inproceedings{EN5,
  title={Towards generative aspect-based sentiment analysis},
  author={Zhang, Wenxuan and Li, Xin and Deng, Yang and Bing, Lidong and Lam, Wai},
  booktitle={Proceedings of the 59th Annual Meeting of the Association for Computational Linguistics and the 11th International Joint Conference on Natural Language Processing (Volume 2: Short Papers)},
  pages={504--510},
  year={2021}
}

@article{mt5,
  title={mT5: A massively multilingual pre-trained text-to-text transformer},
  author={Xue, Linting and Constant, Noah and Roberts, Adam and Kale, Mihir and Al-Rfou, Rami and Siddhant, Aditya and Barua, Aditya and Raffel, Colin},
  journal={arXiv preprint arXiv:2010.11934},
  year={2020}
}

@article{LLMABSA1,
  title={Large language models for aspect-based sentiment analysis},
  author={Simmering, Paul F and Huoviala, Paavo},
  journal={arXiv preprint arXiv:2310.18025},
  year={2023}
}

@article{LLMABSA2,
  title={A Comprehensive Evaluation of Large Language Models on Aspect-Based Sentiment Analysis},
  author={Zhou, Changzhi and others},
  journal={arXiv preprint arXiv:2412.02279},
  year={2024}
}

@inproceedings{LLMABSA3,
  title={Leveraging Large Language Models for Aspect-Based Sentiment Classification Using GPT-4},
  author={Alanazi, Sami and Liu, Xiuwen},
  booktitle={International Journal of Advanced Information Technology},
  volume={14},
  number={4},
  year={2024}
}

@article{LLMABSA4,
  title={Multi-Domain ABSA Conversation Dataset Generation via LLMs for Real-World Evaluation and Model Comparison},
  author={Pandit, Tejul and others},
  journal={arXiv preprint arXiv:2505.24701},
  year={2025}
}

@article{LLMABSA5,
  title={From Annotation to Adaptation: Metrics, Synthetic Data, and Aspect Extraction for Aspect-Based Sentiment Analysis with Large Language Models},
  author={Neveditsin, Nikita and Lingras, Pawan and Mago, Vijay},
  journal={arXiv preprint arXiv:2503.20715},
  year={2025}
}

@inproceedings{ParaPhrase,
  title={Aspect Sentiment Quad Prediction as Paraphrase Generation},
  author={Zhang, Wenxuan and Deng, Yang and Li, Xin and Yuan, Yifei and Bing, Lidong and Lam, Wai},
  booktitle={Proceedings of the 2021 Conference on Empirical Methods in Natural Language Processing},
  pages={9209--9219},
  year={2021}
}

@article{qwen3,
  title={Qwen3 technical report},
  author={Yang, An and Li, Anfeng and Yang, Baosong and Zhang, Beichen and Hui, Binyuan and Zheng, Bo and Yu, Bowen and Gao, Chang and Huang, Chengen and Lv, Chenxu and others},
  journal={arXiv preprint arXiv:2505.09388},
  year={2025}
}

@misc{gpt-5,
  author       = {OpenAI},
  title        = {Introducing GPT-5},
  year         = {2025},
  url          = {https://openai.com/index/introducing-gpt-5/}
}

@inproceedings{Span-ASTE,
  title={Learning Span-Level Interactions for Aspect Sentiment Triplet Extraction},
  author={Xu, Lu and Chia, Yew Ken and Bing, Lidong},
  booktitle={Proceedings of the 59th Annual Meeting of the Association for Computational Linguistics and the 11th International Joint Conference on Natural Language Processing (Volume 1: Long Papers)},
  pages={4755--4766},
  year={2021}
}

@inproceedings{ACOS,
  title={Aspect-Category-Opinion-Sentiment Quadruple Extraction with Implicit Aspects and Opinions},
  author={Cai, Hongjie and Xia, Rui and Yu, Jianfei},
  booktitle={Proceedings of the 59th Annual Meeting of the Association for Computational Linguistics and the 11th International Joint Conference on Natural Language Processing (Volume 1: Long Papers)},
  pages={340--350},
  year={2021}
}

@inproceedings{shengchengshi,
    title = "Self-Consistent Reasoning-based Aspect-Sentiment Quad Prediction with Extract-Then-Assign Strategy",
    author = "Kim, Jieyong  and
      Heo, Ryang  and
      Seo, Yongsik  and
      Kang, SeongKu  and
      Yeo, Jinyoung  and
      Lee, Dongha",
    editor = "Ku, Lun-Wei  and
      Martins, Andre  and
      Srikumar, Vivek",
    booktitle = "Findings of the Association for Computational Linguistics: ACL 2024",
    month = aug,
    year = "2024",
    address = "Bangkok, Thailand",
    publisher = "Association for Computational Linguistics",
    url = "https://aclanthology.org/2024.findings-acl.435/",
    doi = "10.18653/v1/2024.findings-acl.435",
    pages = "7295--7303"
}

@inproceedings{jiegoushengcheng,
    title = "Transition-based Opinion Generation for Aspect-based Sentiment Analysis",
    author = "Ma, Tianlai  and
      Wang, Zhongqing  and
      Zhou, Guodong",
    editor = "Ku, Lun-Wei  and
      Martins, Andre  and
      Srikumar, Vivek",
    booktitle = "Findings of the Association for Computational Linguistics: ACL 2024",
    month = aug,
    year = "2024",
    address = "Bangkok, Thailand",
    publisher = "Association for Computational Linguistics",
    url = "https://aclanthology.org/2024.findings-acl.182/",
    doi = "10.18653/v1/2024.findings-acl.182",
    pages = "3078--3087"
}

@inproceedings{uy1,
    title = "Building Language Models for Morphological Rich Low-Resource Languages using Data from Related Donor Languages: the Case of {U}yghur",
    author = "Abulimiti, Ayimunishagu  and
      Schultz, Tanja",
    editor = "Beermann, Dorothee  and
      Besacier, Laurent  and
      Sakti, Sakriani  and
      Soria, Claudia",
    booktitle = "Proceedings of the 1st Joint Workshop on Spoken Language Technologies for Under-resourced languages (SLTU) and Collaboration and Computing for Under-Resourced Languages (CCURL)",
    month = may,
    year = "2020",
    address = "Marseille, France",
    publisher = "European Language Resources association",
    url = "https://aclanthology.org/2020.sltu-1.38/",
    pages = "271--276",
    language = "eng",
    ISBN = "979-10-95546-35-1"
}

@inproceedings{Turkish,
    title = "Two-level Description of {T}urkish Morphology",
    author = "Oflazer, Kemal",
    editor = "Krauwer, Steven  and
      Moortgat, Michael  and
      des Tombe, Louis",
    booktitle = "Sixth Conference of the {E}uropean Chapter of the Association for Computational Linguistics",
    month = apr,
    year = "1993",
    address = "Utrecht, The Netherlands",
    publisher = "Association for Computational Linguistics",
    url = "https://aclanthology.org/E93-1066/"
}

@inproceedings{nianzhuoyu,
    title = "Why do language models perform worse for morphologically complex languages?",
    author = "Arnett, Catherine  and
      Bergen, Benjamin",
    editor = "Rambow, Owen  and
      Wanner, Leo  and
      Apidianaki, Marianna  and
      Al-Khalifa, Hend  and
      Eugenio, Barbara Di  and
      Schockaert, Steven",
    booktitle = "Proceedings of the 31st International Conference on Computational Linguistics",
    month = jan,
    year = "2025",
    address = "Abu Dhabi, UAE",
    publisher = "Association for Computational Linguistics",
    url = "https://aclanthology.org/2025.coling-main.441/",
    pages = "6607--6623"
}

@inproceedings{xiangguan,
    title = "Aspect-based Sentiment Classification with Aspect-specific Graph Convolutional Networks",
    author = "Zhang, Chen  and
      Li, Qiuchi  and
      Song, Dawei",
    editor = "Inui, Kentaro  and
      Jiang, Jing  and
      Ng, Vincent  and
      Wan, Xiaojun",
    booktitle = "Proceedings of the 2019 Conference on Empirical Methods in Natural Language Processing and the 9th International Joint Conference on Natural Language Processing (EMNLP-IJCNLP)",
    month = nov,
    year = "2019",
    address = "Hong Kong, China",
    publisher = "Association for Computational Linguistics",
    url = "https://aclanthology.org/D19-1464/",
    doi = "10.18653/v1/D19-1464",
    pages = "4568--4578"
}

@inproceedings{xiangguan1,
    title = "Encoding Sentences with Graph Convolutional Networks for Semantic Role Labeling",
    author = "Marcheggiani, Diego  and
      Titov, Ivan",
    editor = "Palmer, Martha  and
      Hwa, Rebecca  and
      Riedel, Sebastian",
    booktitle = "Proceedings of the 2017 Conference on Empirical Methods in Natural Language Processing",
    month = sep,
    year = "2017",
    address = "Copenhagen, Denmark",
    publisher = "Association for Computational Linguistics",
    url = "https://aclanthology.org/D17-1159/",
    doi = "10.18653/v1/D17-1159",
    pages = "1506--1515"
}

@inproceedings{xiangguan2,
    title = "{U}niversal {D}ependencies v2: An Evergrowing Multilingual Treebank Collection",
    author = "Nivre, Joakim  and
      de Marneffe, Marie-Catherine  and
      Ginter, Filip  and
      Haji{\v{c}}, Jan  and
      Manning, Christopher D.  and
      Pyysalo, Sampo  and
      Schuster, Sebastian  and
      Tyers, Francis  and
      Zeman, Daniel",
    editor = "Calzolari, Nicoletta  and
      B{\'e}chet, Fr{\'e}d{\'e}ric  and
      Blache, Philippe  and
      Choukri, Khalid  and
      Cieri, Christopher  and
      Declerck, Thierry  and
      Goggi, Sara  and
      Isahara, Hitoshi  and
      Maegaard, Bente  and
      Mariani, Joseph  and
      Mazo, H{\'e}l{\`e}ne  and
      Moreno, Asuncion  and
      Odijk, Jan  and
      Piperidis, Stelios",
    booktitle = "Proceedings of the Twelfth Language Resources and Evaluation Conference",
    month = may,
    year = "2020",
    address = "Marseille, France",
    publisher = "European Language Resources Association",
    url = "https://aclanthology.org/2020.lrec-1.497/",
    pages = "4034--4043",
    language = "eng",
    ISBN = "979-10-95546-34-4"
}

@inproceedings{xiangguan3,
    title = "{S}tanza: A Python Natural Language Processing Toolkit for Many Human Languages",
    author = "Qi, Peng  and
      Zhang, Yuhao  and
      Zhang, Yuhui  and
      Bolton, Jason  and
      Manning, Christopher D.",
    editor = "Celikyilmaz, Asli  and
      Wen, Tsung-Hsien",
    booktitle = "Proceedings of the 58th Annual Meeting of the Association for Computational Linguistics: System Demonstrations",
    month = jul,
    year = "2020",
    address = "Online",
    publisher = "Association for Computational Linguistics",
    url = "https://aclanthology.org/2020.acl-demos.14/",
    doi = "10.18653/v1/2020.acl-demos.14",
    pages = "101--108"
}

@inproceedings{xiangguan4,
    title = "Syntax-{BERT}: Improving Pre-trained Transformers with Syntax Trees",
    author = "Bai, Jiangang  and
      Wang, Yujing  and
      Chen, Yiren  and
      Yang, Yaming  and
      Bai, Jing  and
      Yu, Jing  and
      Tong, Yunhai",
    editor = "Merlo, Paola  and
      Tiedemann, Jorg  and
      Tsarfaty, Reut",
    booktitle = "Proceedings of the 16th Conference of the European Chapter of the Association for Computational Linguistics: Main Volume",
    month = apr,
    year = "2021",
    address = "Online",
    publisher = "Association for Computational Linguistics",
    url = "https://aclanthology.org/2021.eacl-main.262/",
    doi = "10.18653/v1/2021.eacl-main.262",
    pages = "3011--3020"
}

@inproceedings{xiangguan5,
    title = "Grid Tagging Scheme for Aspect-oriented Fine-grained Opinion Extraction",
    author = "Wu, Zhen  and
      Ying, Chengcan  and
      Zhao, Fei  and
      Fan, Zhifang  and
      Dai, Xinyu  and
      Xia, Rui",
    editor = "Cohn, Trevor  and
      He, Yulan  and
      Liu, Yang",
    booktitle = "Findings of the Association for Computational Linguistics: EMNLP 2020",
    month = nov,
    year = "2020",
    address = "Online",
    publisher = "Association for Computational Linguistics",
    url = "https://aclanthology.org/2020.findings-emnlp.234/",
    doi = "10.18653/v1/2020.findings-emnlp.234",
    pages = "2576--2585"
}

@inproceedings{xiangguan6,
    title = "Fusing Heterogeneous Factors with Triaffine Mechanism for Nested Named Entity Recognition",
    author = "Yuan, Zheng  and
      Tan, Chuanqi  and
      Huang, Songfang  and
      Huang, Fei",
    editor = "Muresan, Smaranda  and
      Nakov, Preslav  and
      Villavicencio, Aline",
    booktitle = "Findings of the Association for Computational Linguistics: ACL 2022",
    month = may,
    year = "2022",
    address = "Dublin, Ireland",
    publisher = "Association for Computational Linguistics",
    url = "https://aclanthology.org/2022.findings-acl.250/",
    doi = "10.18653/v1/2022.findings-acl.250",
    pages = "3174--3186"
}


\end{document}